\documentclass[10pt,twocolumn,letterpaper]{article}

\usepackage{cvpr}              
\definecolor{cvprblue}{rgb}{0.21,0.49,0.74}
\usepackage[pagebackref,breaklinks,colorlinks,allcolors=cvprblue]{hyperref}

\def\paperID{10173} 
\def\confName{CVPR}
\def\confYear{2026}

\title{Image Diffusion Models Exhibit Emergent Temporal Propagation in Videos}

\author{
Youngseo Kim \qquad
Dohyun Kim \qquad
Geonhee Han \qquad
Paul Hongsuck Seo\textsuperscript{$\dagger$}\\[4pt]
Dept. of CSE, Korea University\\
    {\tt\small
    \{\href{mailto:xwsa568@korea.ac.kr}{xwsa568},
    \href{mailto:a12s12@korea.ac.kr}{a12s12},
    \href{mailto:rtrt505@korea.ac.kr}{rtrt505},
    \href{mailto:phseo@korea.ac.kr}{phseo}\}@korea.ac.kr
    }
}

\usepackage{graphicx}
\usepackage{caption}
\usepackage{subcaption}
\usepackage[capitalise]{cleveref}
\usepackage{pifont}
\usepackage{multirow}
\usepackage[table]{xcolor}
\usepackage{array}
\usepackage{needspace}
\usepackage{booktabs}
\usepackage{wrapfig}
\usepackage{pgfplots}
\pgfplotsset{compat=1.18}
\usepgfplotslibrary{groupplots}
\usepackage{comment}

\crefname{figure}{Figure}{Figures}
\crefname{table}{Table}{Tables}
\crefname{section}{Section}{Sections}
\crefname{subsection}{Section}{Sections}
\crefname{subsubsection}{Section}{Sections}
\crefname{equation}{Equation}{Equations}
\crefname{algorithm}{Algorithm}{Algorithms}
\crefname{appendix}{Appendix}{Appendices}
\crefname{theorem}{Theorem}{Theorems}
\crefname{lemma}{Lemma}{Lemmas}
\crefname{proposition}{Proposition}{Propositions}
\crefname{corollary}{Corollary}{Corollaries}
\crefname{definition}{Definition}{Definitions}
\crefname{remark}{Remark}{Remarks}

\newcommand{\drift}{\textsc{Drift}}

\DeclareMathOperator{\argmax}{argmax}

\begin{document}
\maketitle
\renewcommand{\thefootnote}{\fnsymbol{footnote}}
\footnotetext[2]{Corresponding author.}
\begin{abstract}
Image diffusion models, though originally developed for image generation, implicitly capture rich semantic structures that enable various recognition and localization tasks beyond synthesis.
In this work, we investigate their self-attention maps can be reinterpreted as semantic label propagation kernels, providing robust pixel-level correspondences between relevant image regions.
Extending this mechanism across frames yields a temporal propagation kernel that enables zero-shot object tracking via segmentation in videos.
We further demonstrate the effectiveness of test-time optimization strategies—DDIM inversion, textual inversion, and adaptive head weighting—in adapting diffusion features for robust and consistent label propagation.
Building on these findings, we introduce \drift{}, 
a framework for object tracking in videos leveraging a pretrained image diffusion model with SAM-guided mask refinement, achieving state-of-the-art zero-shot performance on standard video object segmentation benchmarks.
\end{abstract}    
\section{Introduction}
Diffusion models~\citep{dpm, ddpm, ddim, song2020score, dit} have emerged as the dominant paradigm in generative modeling, achieving remarkable fidelity and diversity across modalities including images, audio, and video. 
Originally developed for image synthesis~\citep{stablediffusion, dalle2, imagen, sdxl}, they have since inspired a broad range of research beyond generation, driven by the observation that their intermediate representations encode rich semantic structures~\citep{hedlin2023unsupervised, diffusionhyperfeatures, dift}. 
This arises naturally from the denoising process itself: to reconstruct coherent visual content from noisy inputs, the model must implicitly learn semantic correspondences among image regions.

Recent studies have demonstrated that diffusion models encode rich semantic representations, enabling their features to be utilized for visual recognition~\citep{diffseg, diffcut, vidseg} and localization tasks~\citep{diffsegmenter, vd-it, difftracker, smite}. 
In this work, we further investigate whether the capabilities learned by diffusion models from static images can capture temporal dynamics in videos by repurposing self-attention into cross-frame attention, revealing that diffusion features inherently support temporal reasoning beyond spatial recognition. 
We leverage these attention maps as a propagation kernel, enabling pixel-level correspondence and temporally consistent mask evolution without any additional supervision.

We further demonstrate that test-time optimization effectively adapts image diffusion representations for object tracking in videos, enhancing temporal stability and object fidelity.
By jointly applying DDIM inversion for semantically preserved noise, mask-specific textual inversion for object-aware adaptation, and adaptive head weighting to combine complementary attention patterns, we achieve stable and consistent mask propagation across frames.
Building on these findings, we present \drift{}, a framework for object tracking in videos leveraging a pretrained image diffusion model with SAM-guided mask refinement~\citep{sam}, achieving state-of-the-art zero-shot performance on standard video object segmentation benchmarks.

Our contributions can be summarized as follows:  
\begin{itemize}
    \item We investigate that the self-attention in pretrained image diffusion models exhibits emergent temporal propagation, enabling zero-shot object tracking in videos.
    \item We demonstrate that three test-time optimization techniques applied to diffusion models further improve cross-frame label propagation.
    \item We propose \drift{}, a framework for object tracking in videos that demonstrates state-of-the-art zero-shot performance through extensive experiments on four standard video object segmentation benchmarks.
\end{itemize}
\begin{figure*}[t]
    \centering
    \begin{subfigure}{0.18\textwidth}
        \centering
        \includegraphics[width=0.7\linewidth]{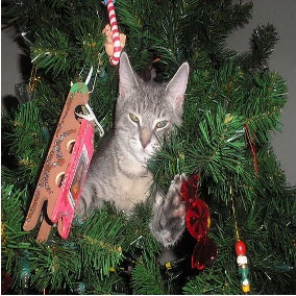}
        \caption{Original Image}
    \end{subfigure}
    \begin{subfigure}{0.18\textwidth}
        \centering
        \includegraphics[width=0.7\linewidth]{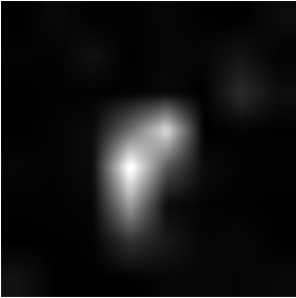}
        \caption{Coarse Map}
    \end{subfigure}
    \begin{subfigure}{0.18\textwidth}
        \centering
        \includegraphics[width=0.7\linewidth]{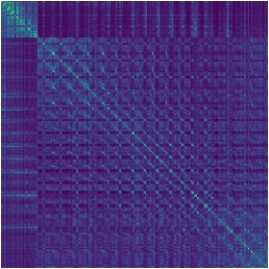}
        \caption{Self-attention Affinity}
    \end{subfigure}
    \begin{subfigure}{0.18\textwidth}
        \centering
        \includegraphics[width=0.7\linewidth]{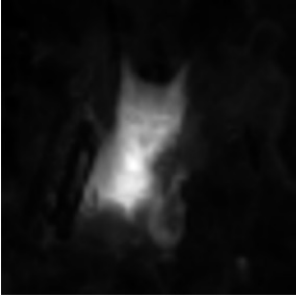}
        \caption{Propagated Mask}
    \end{subfigure}
    \begin{subfigure}{0.18\textwidth}
        \centering
        \includegraphics[width=0.7\linewidth]{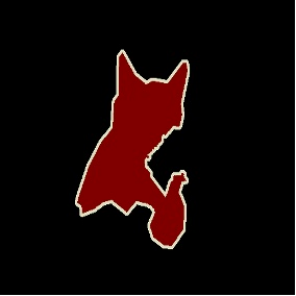}
        \caption{GT Mask}
    \end{subfigure}
    \caption{
        \textbf{Visualization of Label Propagation via Self-Attention from a Text-to-Image Diffusion Models.}
        Given an input image (a), the coarse map (b)—which corresponds to the cross-attention response for the token “cat”—provides approximate object localization based on the text prompt, while the self-attention map (c) captures semantic affinities {across image regions to refine the coarse localization.} Leveraging the self-attention map as a learned label propagation kernel, the coarse map is propagated to yield the final mask (d), which achieves substantially improved spatial precision, closely aligning with the GT mask (e).
    }
    \label{fig:diffsegmenter}
\end{figure*}

\section{Related Work}
\label{sec:relwork}
\noindent\textbf{Vision Tasks Using Image Diffusion Features.} \ \ 
Image diffusion model features have been applied to various vision tasks beyond image generation, including correspondence~\citep{hedlin2023unsupervised, diffusionhyperfeatures, dift} and object detection~\citep{generalized_diffdet}.
For segmentation, many studies have leveraged diffusion models~\citep{diffseg, diffsegmenter, diffcut, diffews, odise}.  
In particular, DiffSeg~\citep{diffseg}, DiffewS~\citep{diffews}, and DiffSegmenter~\citep{diffsegmenter} use attention maps in diffusion model within the image domain to derive or refine spatial masks.
While prior works focus on static-image analysis, our method transfers the representational knowledge of pretrained image diffusion models to the video domain, leveraging diffusion features for object tracking via segmentation and 
achieving temporally consistent mask propagation without task-specific training.

\noindent\textbf{Video Diffusion Models.} \ \ 
Video diffusion models are trained on videos and jointly model multiple frames in a sequence, using spatiotemporal attention to produce temporally coherent motion and appearance~\citep{videoldm, animatediff, videocrafter, stablevideodiffusion}.
Analogous to how image diffusion features are used in image tasks, video diffusion models are used in video applications such as point tracking~\citep{nam2025emergent} and video editing~\citep{qi2023fatezero, wu2023tune, flatten, feng2024wave} by extracting multi-frame video features.
In contrast, we focus on pretrained image diffusion models that have never seen video data, yet we demonstrate that they can still capture meaningful temporal dynamics in videos.

\noindent\textbf{Diffusion-based Object Tracking.} \ \ 
Recent works have explored leveraging diffusion priors for video understanding and tracking.
VD-IT~\citep{vd-it} simply extends prior diffusion-feature–based approaches to the video domain by training an additional segmentation head on top of video-diffusion features.
In contrast, we show that pretrained image diffusion models can perform object tracking through their inherent self-attention, without video data and task-specific supervision.
Diff-Tracker~\citep{difftracker} localizes objects over time using cross-attention maps from a pretrained text-to-image diffusion model, where cross-attention conditioned on textual prompts delineates object regions within each frame. 
To capture motion beyond these spatial correspondences, it introduces a learned motion encoder and an online prompt updater that dynamically adapt the attention to temporal changes. 
SMITE~\citep{smite} also relies on cross-attention features tied to learned text embeddings and fine-tunes those layers for video adaptation, while integrating an external point tracker~\citep{cotracker} to enforce temporal coherence across frames. 
Both approaches depend on spatially grounded cross-attention and require auxiliary components—either learned motion encoders or explicit trackers—to model temporal dynamics. 
In contrast, we show that the self-attention layers of a pretrained text-to-image diffusion model inherently encode transferable motion cues across frames, enabling temporal label propagation without any task-specific training, or external tracking mechanism.

\begin{figure*}[t]
    \centering
        \begin{subfigure}{0.23\textwidth}
            \centering
            \includegraphics[width=1.0\linewidth]{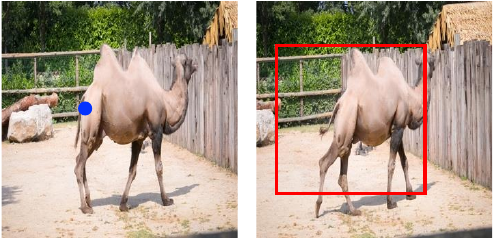}
            \par
            {\scriptsize Frame $t \rightarrow t'$}
            \vspace{8pt}
            \caption{Input Frames}
        \end{subfigure}
        \hspace{6pt}
        \begin{subfigure}{0.23\textwidth}
            \centering
            \includegraphics[width=0.78\linewidth]{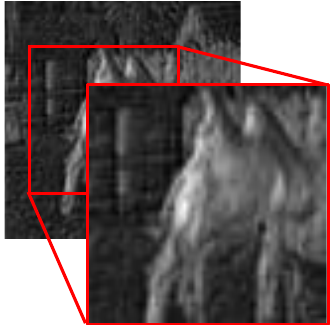}
            \caption{Cosine Similarity}
        \end{subfigure}
        \begin{subfigure}{0.23\textwidth}
            \centering
            \includegraphics[width=0.78\linewidth]{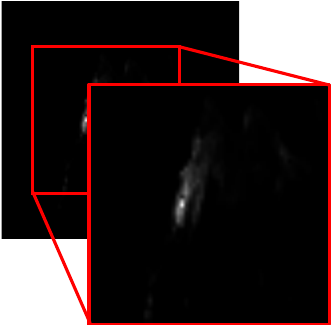}
            \caption{Aggregated Self-attention}
        \end{subfigure}
        \hspace{6pt}
        \begin{subfigure}{0.23\textwidth}
            \centering
            \includegraphics[width=1.0\linewidth]{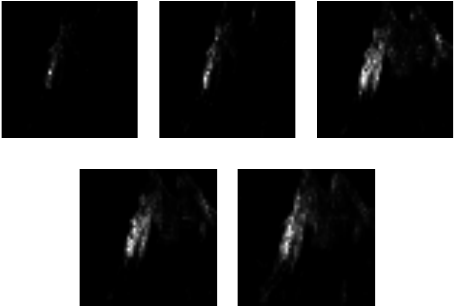}
            \vspace{1pt}
            \caption{Per-head Self-attention}
        \end{subfigure}
    \caption{
        \textbf{Comparison of Cosine Similarity vs. Self-attention for Label Propagation.}
        (a) The blue dot in the frame $t$ is propagated to frame $t{'}$. 
        (b) Cosine similarity produces dispersed activations scattered across unrelated regions. 
        (c) The aggregated self-attention map, in contrast, focuses sharply on the corresponding object region. 
        (d) Individual attention heads exhibit complementary but distinct patterns, highlighting the diverse semantic relationships captured by multi-head self-attention.
    }
    \label{fig:cos_vs_attn}
\end{figure*}

\section{Method}

In this work, our primary goal is to show that pretrained text-to-image diffusion models can be repurposed as object trackers without task-specific finetuning.
The key insight is that the pairwise query–key interactions in diffusion self-attention naturally support label propagation across frames, which forms the foundation of object tracking by segmentation. We further identify the self-attention mechanism as the central component enabling this propagation.

\subsection{Temporal Label Propagation via Self-Attention}
\label{sec:cross_frame}

Recent study~\citep{diffsegmenter} demonstrates that pretrained diffusion models can perform open-voca segmentation without task-specific training. This ability arises from the cross-attention layers, which align text and visual tokens and thereby highlight visual regions corresponding to text-specified classes.
However, the spatial maps produced by cross-attention alone are typically coarse. Accuracy improves only when these maps are multiplied by self-attention maps, as shown in~\cref{fig:diffsegmenter}. This indicates that self-attention serves as a label propagation kernel: activations at one pixel can propagate to other pixels with similar semantics, refining a coarse mask into a more detailed segmentation. In this sense, the self-attention map functions as a learned mechanism for semantic label propagation.
Formally, the self-attention map at layer $l$ is defined as:
\begin{align}
A^{(l,h)}_\mathrm{self} = 
\text{softmax}\!\left(
    \frac{Q^{(l,h)} \cdot {K^{(l,h)}}^\top}{\sqrt{d}}
\right),
\label{eq:self_attn}
\end{align}
where $Q^{(l,h)},K^{(l,h)} \in \mathbb{R}^{N\times d}$ are the query and key matrices of head $h$ in layer $l$, 
$N=H\times W$ is the number of spatial locations, and $d$ is the dimension per head.
Averaging over all layers and heads yields $A_\mathrm{self}$, which encodes semantic affinities between pixel pairs 
and serves as a propagation kernel that spreads coarse activations into fine-grained segmentation masks.

Building on this interpretation, we extend label propagation from the spatial domain of a single image to the temporal domain of a video. Given two consecutive frames $I_{t-1}$ and $I_t$, we compute a cross-frame attention map that measures similarities between features across frames:
\begin{align}
\bar{A}_{t,t-1} = \sum_{l \in \mathcal{L}}
    \left( \sum_{h=1}^{H}
        w^{(l,h)}\text{softmax}\!\left(\frac{Q^{(l,h)}_{t} \cdot {K^{(l,h)}_{t-1}}^\top}{\sqrt{d}}\right)
    \right),
\label{eq:cross_frame}
\end{align}
where $Q^{(l,h)}_{t}$ and $K^{(l,h)}_{t-1}$ are the query and key matrices from head $h$ of layer $l$ in frames $t$ and $t-1$, respectively, and $w^{(l,h)}$ is the weight assigned to each head (by default $w^{(l,h)} = \tfrac{1}{|\mathcal{L}|\times H}$). Each row of $\bar{A}_{t,t-1}$ defines how the label at a pixel in frame $t$ should aggregate information from frame $t-1$. The propagated mask is then updated as $\hat{M}_t = \bar{A}_{t,t-1} \hat{M}_{t-1}.$
In this way, the image diffusion model’s self-attention is repurposed as a cross-frame label propagation kernel, enabling masks specified in the first frame to be consistently propagated through the video in a zero-shot manner.

\subsection{Raw Feature Similarity vs. Self-Attention}
\label{sec:cos_vs_attn}
Many prior approaches~\citep{stc, dino} often rely on cosine similarity of learned features for label propagation, and several recent studies have explored using pretrained diffusion features for this purpose~\citep{diffcut, dift}. In these works, the raw diffusion features are extracted and directly used as visual representations. However, we find that relying on raw diffusion features and measuring pairwise cosine similarity overlooks reusable pretrained knowledge embedded in the self-attention layers of diffusion models. 

A self-attention map inherently captures feature similarity, but unlike cosine similarity, it does so after learned query and key projections that act as filters, preserving certain salient aspects for similarity. 
Moreover, multi-head self-attention incorporates multiple such projections, enabling the model to capture diverse semantic relationships and produce more robust similarity maps.
In contrast, raw diffusion features—being optimized for image generation—may encode aspects that are irrelevant to semantic similarity. As a result, cosine similarity over these features often produces noisy maps dispersed across unrelated regions.
The example in \cref{fig:cos_vs_attn} illustrates this and supports our arguments.
Given two input frames (a), the cosine similarity of features (b) propagates the blue point in frame $t$ beyond the relevant region in frame $t'$, dispersing it across the entire image and even into irrelevant background areas.
As shown in (d), the multiple heads of self-attention each highlight different but still relevant regions, providing complementary views of semantic similarity.
By aggregating these heads in (c), the model effectively emphasizes the relevant region while suppressing spurious propagation.

\subsection{Test-Time Optimization for Label Propagation}
\label{sec:techniques}
Although diffusion self-attention inherently captures semantic correspondences that enable label propagation, its raw form is often insufficient for reliable mask propagation across frames. 
We therefore investigate how to strengthen its propagation capability through three complementary test-time optimization techniques for diffusion models—DDIM inversion, mask-specific textual inversion, and adaptive head weighting—which together yield more accurate and object-aware masks.

\subsubsection{Inversion for Semantics Preserving Noise}
\label{sec:ddim_inversion}
Diffusion models are inherently designed to take noisy images as input, but excessive or insufficient noise can distort semantics.
At large timesteps, the input noise representation is dominated by the noise component and loses original semantic information,
while noise-free inputs fully preserve semantic information, yet the diffusion model fails to leverage it effectively since its denoising objective offers little incentive to capture semantics.
As a result, the final propagation quality can be highly sensitive to the chosen noise level.
To reduce this sensitivity, we employ DDIM inversion~\citep{ddim,diffusionbeatgan}, which perturbs the input image using model-predicted noise instead of random Gaussian noise. 
This process aligns the noise representation with the model’s learned semantic manifold, yielding more stable and semantically consistent features across different timesteps.

While this technique has been widely adapted in video editing~\citep{flatten, feng2024wave, wu2023tune, qi2023fatezero}, primarily to ensure faithful reconstruction of unedited content, our study demonstrates a distinct advantage for label propagation.
Specifically, we show that DDIM inversion enables content-specific representations from a diffusion model, producing attention maps particularly well-suited for propagating labels across frames.

\subsubsection{Mask-Specific Prompts via Textual Inversion}
\label{sec:text_inversion}

To exploit cross-frame attention maps from a text-to-image diffusion model for label propagation, the model requires a text prompt as input—something not naturally available for object tracking. 
A na\"ive solution is to use a \texttt{null} prompt, but such prompts fail to capture information about the target object and its visual context. 
To address this, we adopt textual inversion~\citep{textual_inversion}, tuning a set of learnable text tokens specifically for mask propagation.
Leveraging the fact that the GT mask in the initial frame at $t{=}0$ is available, we learn textual embeddings that encourage the model to reproduce the GT mask after propagating it within the same frame, thereby enforcing self-consistent label propagation.
To this end, we obtain a self-attention map that serves as a label-propagation kernel within the initial frame by feeding the frame as both query and key to \cref{eq:cross_frame}, while conditioning the computation on the learnable text embeddings $\theta$.
This yields an aggregated attention map $\bar{A}_{0,0}(\theta)$ through which the GT mask $M_0$ is propagated as $\hat{M}_0(\theta)=\bar{A}_{0,0}(\theta)\cdot M_0$.
Since propagation occurs within the same image, $\hat{M}_0(\theta)$ should ideally reconstruct the original mask $M_0$; that is, the propagation kernel $\bar{A}_{0,0}(\theta)$ should restrict label propagation precisely to the target object region.
Reflecting this principle, we design the following loss to optimize $\theta$:
\begin{equation}
\mathcal{J}(\theta)= \frac{1}{N} \sum_{i=1}^{N} \mathrm{BCE}\left( \hat{M}_0^{(i)}(\theta), M_0^{(i)} \right),
\label{eq:text_inversion}
\end{equation}
where $\mathrm{BCE}$ denotes binary cross-entropy, and $M_0^{(i)}$ and $\hat{M}_0^{(i)}(\theta)$ represent the values at the $i$-th spatial location in the GT and predicted masks, respectively.

The textual inversion technique was originally proposed to learn tokens that are semantically aligned with visual concepts in specific image regions~\citep{textual_inversion, dreambooth}.
In related tasks~\citep{difftracker, smite}, these tokens are optimized to produce cross-attention maps that localize GT segmentation masks within a single frame, without modeling temporal relationships across frames.
As a result, such methods require additional modules to account for temporal dynamics.
In contrast, we focus on self-attention maps—not as final outputs, but as propagation kernels that transfer a given mask across frames, thereby directly modeling temporal dynamics in videos.
Under this formulation, the textual inversion process can be interpreted as learning an object-specific propagation kernel rather than performing object localization.
Given this distinction, we observe that the learned embeddings occupy markedly different manifolds from those representing visual concepts captured by standard text-token embeddings.
We further analyze these learned embeddings in \cref{sec:analyses}.

\subsubsection{Adaptive Weighting of Multi-Head Attention}
\label{sec:head_weight}
As discussed in \cref{sec:cos_vs_attn}, diffusion models employ multi-head self-attention, with each head capturing different semantic correspondences.
Since some heads are more informative than others, we replace uniform averaging with optimized head-specific weights $w^{(l,h)}{\in}[0,1]$, constrained such that 
$\sum_{(l,h)\in \mathcal{L}\times [H]} w^{(l,h)}$.
To this end, we perform test-time optimization, updating these weights jointly with the mask-specific text embeddings $\theta$ by minimizing the loss in \cref{eq:text_inversion}.
This allows informative heads to receive higher weights while less useful ones are down-weighted.
With this refinement, the final attention map becomes a weighted aggregation of all heads, as defined in \cref{eq:cross_frame}, which can improve segmentation quality by emphasizing heads that capture stronger semantic correspondences.

\section{Diffusion-Based Object Tracking Model}
\label{sec:drift}

Building on the temporal label propagation capability of pretrained image diffusion models, we introduce Diffusion-based Region Inference with cross-Frame attention for Tracking (\drift{}), which achieves state-of-the-art performance in zero-shot object tracking via segmentation in videos.
Our approach combines a pretrained text-to-image diffusion model with the Segment Anything Model (SAM)~\citep{sam} for mask refinement.

\subsection{Zero-Shot Object Tracking via Segmentation}
\label{sec:task_def}
The proposed method tackles object tracking via segmentation in a zero-shot setting.
Given a video, the input is an accurate mask of the target object in the first frame, and the goal is to generate precise segmentation masks that trace the object throughout the subsequent frames.
In prior work~\citep{davis2017}, this task is often described as semi-supervised video object segmentation, since the initial mask serves as a form of supervision.
In contrast, we adopt the term object tracking via segmentation to avoid confusion with our zero-shot setup, where no task-specific training data are used.
Formally, a video of $T+1$ frames is denoted as $\mathcal{V} = \{I_0, I_1, \dots, I_T\}$ with $I_t \in \mathbb{R}^{H \times W \times 3}$, and the provided first-frame mask is $M_0 \in \{0,1,\dots,\mathcal{O}\}^{H \times W}$ for $\mathcal{O}$ objects and background (label 0). 
The goal is to predict masks $M_t$ for $t=1,\dots,T$.
Our method, \drift{}, addresses this task in a zero-shot manner by leveraging the inherent label propagation capability of pretrained text-to-image diffusion models discussed in the previous section.

\subsection{Multi-Frame Label Propagation}
\label{sec:multiframe}

To improve robustness in object tracking, we perform label propagation across multiple preceding frames to improve stability, motivated by prior studies~\citep{wang2019learning, li2016unsupervised, lai2019self, stc, dino}.
At frame $t$, we define the set of indices for reference frames as $\mathcal{S}_t = \{0, t{-}1, \dots, t{-}S\}$, which includes $S$ preceding frames and the initial frame.
The final mask is then obtained by aggregating propagated masks as
$\hat{M}_t = \sum_{s \in \mathcal{S}_t} \bar{A}_{t,s}\hat{M}_{s}$.
For stability, each attention map is spatially constrained with a radius-$r$ mask and sparsified by retaining the top-k scores, following~\citep{stc, dino}.
In the multi-object setting~\citep{davis2017}, object and background masks are propagated independently, and the final segmentation is obtained by a pixel-wise argmax, 
$M_t = \argmax_{o \in \{0, \dots, \mathcal{O}\}} \hat{M}_t^{(o)}$.

\begin{figure}[t]
    \centering
    \begin{tikzpicture}
        \begin{axis}[
            width=\columnwidth,
            height=0.58\columnwidth,
            xlabel={Frame $t$},
            ylabel={$\mathcal{J}\&\mathcal{F}_\mathrm{m}$},
            label style={font=\scriptsize},
            xmin=1, xmax=32,
            ymin=30, ymax=95,
            xtick={0,5,10,15,20,25,30},
            ytick={30,50,...,90},
            tick label style={font=\scriptsize},
            grid=both,
            legend style={font=\scriptsize, at={(0.5,1.05)}, anchor=south, legend columns=2, draw=none},
        ]
        
        \addplot+[mark=o, mark size=1pt, color=orange, thick] coordinates {
        (1,40.4)(2,39.2)(3,39.2)(4,38.6)(5,37.7)(6,38.3)(7,37.4)(8,37.7)
        (9,37.3)(10,37.5)(11,36.8)(12,35.9)(13,34.5)(14,34.8)(15,34.4)(16,33.9)
        (17,34.1)(18,33.6)(19,33.6)(20,35.0)(21,37.5)(22,36.5)(23,37.0)(24,37.0)
        (25,36.9)(26,35.7)(27,35.7)(28,35.9)(29,35.5)(30,35.6)(31,35.4)(32,35.0)
        };
        \addlegendentry{Cosine Similarity}
        
        \addplot+[mark=square, mark size=1pt, color=blue, thick] coordinates {
        (1,85.9)(2,85.0)(3,83.8)(4,83.0)(5,82.7)(6,81.9)(7,79.8)(8,78.2)
        (9,77.5)(10,77.1)(11,75.7)(12,73.7)(13,72.3)(14,70.5)(15,70.1)(16,69.3)
        (17,68.9)(18,68.3)(19,68.0)(20,66.7)(21,66.6)(22,66.5)(23,65.2)(24,65.2)
        (25,65.1)(26,65.4)(27,65.0)(28,63.9)(29,63.6)(30,62.7)(31,62.4)(32,61.4)
        };
        \addlegendentry{Self-attention}
        
        \node[inner sep=0pt, anchor=south west] (img1) at (axis cs:1.5,53)
            {\includegraphics[width=1.2cm]{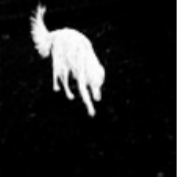}};
        \draw[blue, -stealth] (img1.north) -- (axis cs:2,85.0);
        
        \node[inner sep=0pt, anchor=north west] (img2) at (axis cs:26,92)
            {\includegraphics[width=1.2cm]{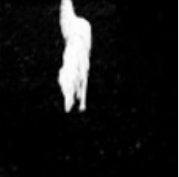}};
        \draw[blue, -stealth] (img2.south) -- (axis cs:26,65.4);
        
        \node[inner sep=0pt, anchor=north east] (img3) at (axis cs:14,65)
            {\includegraphics[width=1.2cm]{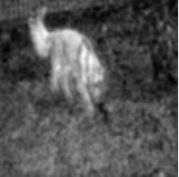}};
        \draw[orange, -stealth] (img3.west) -- (axis cs:2,39.2);
        
        \node[inner sep=0pt, anchor=north east] (img4) at (axis cs:22,62)
            {\includegraphics[width=1.2cm]{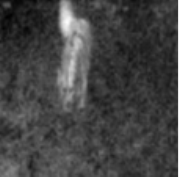}};
        \draw[orange, -stealth] (img4.east) -- (axis cs:26,35.7);
        
        \end{axis}
    \end{tikzpicture}
    \caption{\textbf{Comparison of Per-frame $\mathcal{J}\&\mathcal{F}_\mathrm{m}$ between Self-attention and Cosine-similarity Affinity Maps on DAVIS 2017.}
    Corresponding mask visualizations highlight that cosine-similarity produces noisy and dispersed affinities, whereas self-attention yields spatially precise and temporally consistent masks.
    }
    \label{fig:per-frame}
\end{figure}
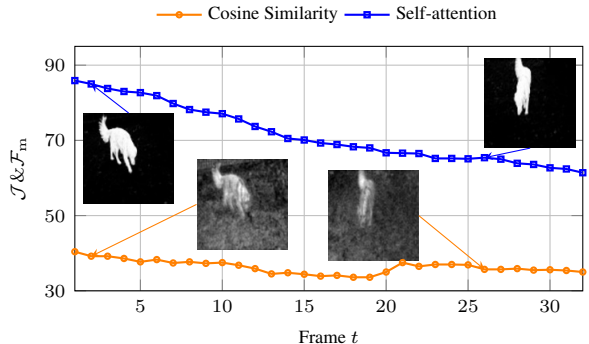

\subsection{Mask Refinement with SAM}
\label{sec:mask_refine}

While the proposed cross-frame attention mechanism already provides a strong prior for object segmentation in a zero-shot manner, the resulting masks may still lack fine-grained detail, as the model is never trained with segmentation supervision.
To enhance mask quality, we further integrate SAM to refine the predicted masks.
To do this, we treat each soft mask prediction $\hat{M}_t^{(o)}$ as a spatial probability distribution by normalizing it so that the pixel values sum to one. 
This normalized distribution serves as a strong prior for the target object's location and shape. 
Based on this, we sample $p$ sets of $n$ point prompts from the distribution and obtain $p$ candidate masks from SAM.
For each of these masks, we compute the IoU with the original predicted mask $\hat{M}_t^{(o)}$ and select the one with the highest score. Details about IoU computation are in Supp. Mat.
Finally, we extract the logits associated with the selected SAM mask and apply the multi-object prediction procedure described earlier to finalize the segmentation.

\section{Experiments}
\subsection{Evaluation Datasets \& Metrics}
\label{sec:settings}

We evaluate our method on four widely used semi-supervised video object segmentation benchmarks:
DAVIS-2016~\cite{davis2016}, DAVIS-2017~\cite{davis2017}, YouTube-VOS 2018~\cite{ytvos2018}, 
and Long Videos~\cite{longvideos}.
Detailed dataset statistics are provided in Supp. Mat. 
We report region similarity $\mathcal{J}_\mathrm{m}$, contour accuracy $\mathcal{F}_\mathrm{m}$, 
and their average $\mathcal{J}\&\mathcal{F}_\mathrm{m}$. 
The $\mathcal{J}_\mathrm{m}$ is the Jaccard index, or intersection-over-union (IoU), 
which measures the overlap between predicted and GT masks, averaged over all annotated objects and frames. 
The $\mathcal{F}_\mathrm{m}$ is the boundary F-measure, computed as the harmonic mean of 
boundary precision and recall, also averaged across objects and frames. 
Finally, $\mathcal{J}\&\mathcal{F}_\mathrm{m}$ is defined as the average of $\mathcal{J}_\mathrm{m}$ and $\mathcal{F}_\mathrm{m}$, 
providing an overall indicator of segmentation quality.
Implementation details are provided in Supp. Mat.

\subsection{Analyses of Cross-Frame Label Propagation}
\label{sec:analyses}

\begin{figure}[t]
\centering
    \begin{tikzpicture}
        \begin{axis}[
            width=0.85\columnwidth,
            height=0.5\columnwidth,
            xlabel={Timestep $\tau$},
            ylabel={$\mathcal{J}\&\mathcal{F}_\mathrm{m}$},
            label style={font=\scriptsize},
            xmin=0, xmax=210,
            ymin=55, ymax=59,
            xtick={0,40,80,120,160,200},
            ytick={55,56,57,58,59},
            tick label style={font=\scriptsize},
            legend style={at={(0.5,1.05)}, font=\scriptsize, anchor=south, legend columns=2, draw=none},
            grid=both,
        ]
        
        \addplot[
            color=orange,
            mark=o,
            thick,
            mark size=1pt
        ] coordinates {
            (1,56.8) (21,57.4) (41,57.4) (61,57.2) (81,57.0)
            (101,56.7) (121,56.4) (141,56.1) (161,55.8) (181,55.5) (201,55.2)
        };
        \addlegendentry{Random}
        
        \addplot[
            color=blue,
            mark=square,
            thick,
            mark size=1pt
        ] coordinates {
            (21,57.5) (41,57.8) (61,57.9) (81,58.0) (101,58.0)
            (121,58.0) (141,57.9) (161,57.8) (181,57.5) (201,56.9)
        };
        \addlegendentry{Inversion}
        
        \end{axis}
    \end{tikzpicture}
    \caption{\textbf{Comparison of $\mathcal{J}\&\mathcal{F}_\mathrm{m}$ between Random Noise Injection and DDIM Inversion across Diffusion Timesteps $\tau$ on DAVIS 2017.}
    DDIM inversion (blue) achieves higher peak performance and maintains stable segmentation quality across timesteps, whereas random noise injection (orange) rapidly degrades after its early peak due to semantic washout at large $\tau$.
    }
    \label{fig:timestep}
\end{figure}
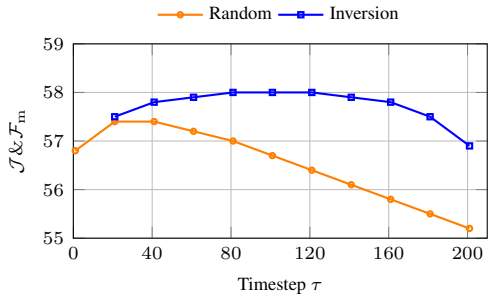

\noindent \textbf{Self-Attention vs. Cosine Similarity} \ \ 
\cref{fig:per-frame} compares cross-frame affinity maps obtained from (i) the self-attention layers of the diffusion model (ours) and (ii) cosine similarity scores computed directly from raw diffusion features following the existing method~\citep{dift}. 
Using these maps, we propagate the initial GT mask to each subsequent frame, and evaluate the per-frame $\mathcal{J}\&\mathcal{F}_\mathrm{m}$ scores on the DAVIS validation set~\citep{davis2017}.
The results reveal a clear performance gap between the cosine-similarity baseline and our self-attention–based affinity maps. 
Cosine similarity, which directly compares raw features, is easily influenced by feature components unrelated to the target object. 
This results in low $\mathcal{J}\&\mathcal{F}_\mathrm{m}$ scores (orange line), caused by dispersed similarity maps, as illustrated in the example propagated masks
—even for frames near the initial mask where appearance changes are minimal.
By contrast, our method exploits the learned projections in the self-attention layers, enabling similarity estimation along semantically meaningful dimensions.
The multi-head design further enriches the label-propagation kernel by capturing multiple complementary aspects of similarity.
Consequently, cross-frame attention maps yield nearly twice the propagation performance of raw-feature cosine similarity across timesteps (blue line), with examples of accurately propagated masks shown alongside.

\begin{table}[t]
\centering
\caption{\textbf{Comparison of Different Prompt Types on DAVIS 2017.}
    Null, class-name, BLIP-2–generated caption, and learned embedding prompts are evaluated, with all results reported in terms of $\mathcal{J}\&\mathcal{F}_\mathrm{m}$, $\mathcal{J}_\mathrm{m}$, and $\mathcal{F}_\mathrm{m}$.
    }
\scalebox{0.9}{
\begin{tabular}{l ccc}
    \toprule
    \toprule
    Prompt type & $\mathcal{J}\&\mathcal{F}_\mathrm{m}$ & $\mathcal{J}_\mathrm{m}$ & $\mathcal{F}_\mathrm{m}$ \\
    \midrule
    Null & 71.8 & 67.9 & 75.6\\
    Class & 71.9 & 68.0 & 75.8\\
    Caption & 72.0 & 68.2 & 75.8\\
    Learned & \textbf{74.5} & \textbf{70.3} & \textbf{78.6}\\
    \bottomrule
    \bottomrule
\end{tabular}
}
\vspace{-1em}
\label{tab:prompts}
\end{table}

\noindent \textbf{Effects of Diffusion Timesteps and DDIM Inversion} \ \ 
We analyze the effect of diffusion timesteps and DDIM inversion on mask propagation performance. 
\cref{fig:timestep} illustrates $\mathcal{J}\&\mathcal{F}_\mathrm{m}$ across different timesteps. 
When injecting white noise, maximum performance is attained at timestep~21 (57.4\%) and then quickly degrades as the timestep increases, reflecting the well-known trade-off~\citep{dift, diffsegmenter} that excessive noise at large timesteps washes out original semantics, while at step~1 the nearly noise-free latents show lower $\mathcal{J}\&\mathcal{F}_\mathrm{m}$ than at step~21. 
In contrast, DDIM inversion—which perturbs latents with model-predicted noise and thus initializes from a model-aligned representation—reaches a higher peak at step~81 (58.0\%) and remains above the random noise curve at all evaluated timesteps. 
This consistent advantage supports our premise that DDIM inversion preserves semantic information more faithfully than the standard random noise injection, yielding more reliable features for mask propagation across a broad range of diffusion timesteps.

\noindent \textbf{Effects of Textual Inversion} \ \ 
In \cref{tab:prompts}, we compare four prompt types: a null prompt (Null), human-annotated object class names such as `dog' or `person' (Class), BLIP-2–generated object-specific captions (Caption; detailed in Supp. Mat.), and learned embeddings obtained through textual inversion~\citep{textual_inversion} (Learned).
While Null simply uses an empty prompt as the embedding, it already provides a strong baseline of 71.8\% in $\mathcal{J\&F}_\mathrm{m}$.
Supplying text prompts semantically aligned with the target object (Class and Caption) yields only marginal improvements, with absolute gains of 0.1\% and 0.2\%.
In contrast, learning text prompts via textual inversion with our propagation loss achieves a substantial improvement of 3.8\%.
These results indicate that prompts aligned with object semantics—such as class names or captions—are not the key drivers of performance in this task, despite it is commonly assumed.
Instead, embeddings learned for the target task of label propagation yield greater improvements, suggesting that propagation quality can be further enhanced through the task-driven test-time optimization.
However, we observed that such textual inversion has a limitation under cosine similarity, and further discussion is available in Supp. Mat.

\begin{figure}[t]
\centering
    \includegraphics[width=\linewidth]{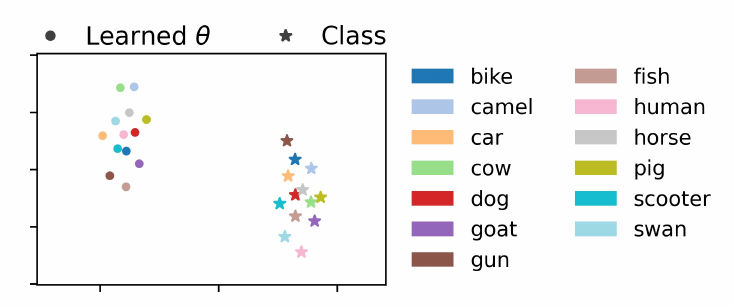}
    \caption{
        \textbf{Visualization of t-SNE Embeddings.}
        Comparison between learned embeddings and class name embeddings, which shows that the learned embeddings form distinct clusters separated from class name embeddings.
    }
    \label{fig:tsne}
\end{figure}

\begin{table}[t]
    \centering
    \caption{
        \textbf{Effect of Attention Head Weighting on DAVIS 2017.}
        $\mathcal{J}\&\mathcal{F}_\mathrm{m}$, $\mathcal{J}_\mathrm{m}$, and $\mathcal{F}_\mathrm{m}$ scores are reported under uniform and learned attention head weighting, evaluated both without and with DDIM inversion (DI) and textual inversion (TI).
    }
    \scalebox{0.9}{
        \begin{tabular}{lcccc}
            \toprule
            \toprule
            Method & $w^{(l,h)}$ & $\mathcal{J}\&\mathcal{F}_\mathrm{m}$ & $\mathcal{J}_\mathrm{m}$ & $\mathcal{F}_\mathrm{m}$ \\
            \midrule
            \multirow{2}{*}{Baseline} 
              & Uniform & 71.1 & 67.0 & 75.1\\
              & \textbf{Learned} & \textbf{71.3} & \textbf{67.2} & \textbf{75.3}\\
            \midrule
            \multirow{2}{*}{+DI+TI} 
              & Uniform & 74.5 & 70.3 & 78.6\\
              & \textbf{Learned} & \textbf{74.8} & \textbf{70.7} & \textbf{78.9}\\
            \bottomrule
            \bottomrule
        \end{tabular}
    }
    \vspace{-1em}
    \label{tab:abl_head}
\end{table}
\begin{table*}[t]
\centering
\caption{
\textbf{Quantitative Comparisons to SOTA Methods.}
Results are reported on~\cite{davis2016,davis2017,ytvos2018,longvideos}.
The subscripts $s$ and $u$ on YT-VOS 2018 indicate seen and unseen categories.
\textit{Zero-shot} methods do not use segmentation annotations during training.
Methods in \textit{zero-shot with image segmentation annotations} section utilizes models that are trained on large image segmentation datasets.
\textit{Fully supervised} models are trained on video segmentation datasets and shown for reference.
$^\dagger$SAM-PT is evaluated using CoTracker~\cite{cotracker}, which is pretrained on a video dataset for dense point tracking.
}
\setlength{\tabcolsep}{4pt}
\scalebox{0.9}{
\begin{tabular}{l@{\hspace{5pt}} ccc@{\hspace{10pt}} ccc@{\hspace{10pt}} ccccc@{\hspace{10pt}} ccc}
\toprule
\toprule
 & \multicolumn{3}{c}{DAVIS 2016}  
 & \multicolumn{3}{c}{DAVIS 2017}  
 & \multicolumn{5}{c}{YT‐VOS 2018}  
 & \multicolumn{3}{c}{Long Videos} \\
\cmidrule(lr{\dimexpr 2\tabcolsep+5pt}){2-4}
\cmidrule(lr{\dimexpr 2\tabcolsep+5pt}){5-7}
\cmidrule(lr{\dimexpr 2\tabcolsep+5pt}){8-12}
\cmidrule(lr){13-15}
Method 
 & $\mathcal{\scriptstyle J}\,\!\text{\scriptsize\&}\!\,\mathcal{\scriptstyle F}_{\mathrm{m}}$
 & $\mathcal{\scriptstyle J}_{\mathrm{m}}$ 
 & $\mathcal{\scriptstyle F}_{\mathrm{m}}$
 & $\mathcal{\scriptstyle J}\,\!\text{\scriptsize\&}\!\,\mathcal{\scriptstyle F}_{\mathrm{m}}$
 & $\mathcal{\scriptstyle J}_{\mathrm{m}}$ 
 & $\mathcal{\scriptstyle F}_{\mathrm{m}}$
 & $\mathcal{\scriptstyle J}\,\!\text{\scriptsize\&}\!\,\mathcal{\scriptstyle F}_{\mathrm{m}}$ 
 & $\mathcal{\scriptstyle J}_{\mathrm{s}}$ 
 & $\mathcal{\scriptstyle F}_{\mathrm{s}}$ 
 & $\mathcal{\scriptstyle J}_{\mathrm{u}}$ 
 & $\mathcal{\scriptstyle F}_{\mathrm{u}}$
 & $\mathcal{\scriptstyle J}\,\!\text{\scriptsize\&}\!\,\mathcal{\scriptstyle F}_{\mathrm{m}}$
 & $\mathcal{\scriptstyle J}_{\mathrm{m}}$ 
 & $\mathcal{\scriptstyle F}_{\mathrm{m}}$ \\
\midrule
\multicolumn{15}{>{\columncolor{gray!20}}l}{\textit{Zero-shot}} \\

STC~\cite{stc}            & 74.5 & 74.7 & 74.4 & 67.6 & 64.8 & 70.2 & 65.5 & 66.0 & 67.1 & 59.8 & 69.2 & 26.2 & 26.4 & 26.0 \\
DIFT~\cite{dift}          & - & - & - & 70.0 & 67.4 & 72.5 & - & - & - & - & - & - & - & - \\
DINO~\cite{dino}          & 81.2 & 80.4 & 81.9 & 71.4 & 67.9 & 74.9 & 62.9 & 64.5 & 67.7 & 53.9 & 65.7 & 45.9 & 45.9 & 45.9 \\
\textbf{DRIFT(Ours)}      & \textbf{85.0} & \textbf{83.7} & \textbf{86.3} & \textbf{74.8} & \textbf{70.7} & \textbf{78.9} & \textbf{68.5} & \textbf{68.1} & \textbf{72.1} & \textbf{61.5} & \textbf{72.4} & \textbf{48.3} & \textbf{48.5} & \textbf{48.1} \\
\midrule
\multicolumn{15}{>{\columncolor{gray!20}}l}{\textit{Zero-shot with Image Segmentation Annotations}} \\
SegGPT~\cite{seggpt}      & 82.3 & 81.8 & 82.8 & 75.6 & 72.5 & 78.6 & 74.7 & \textbf{75.1} & \textbf{80.2} & 67.4 & 75.9 & 21.2 & 18.6 & 23.7 \\
SAM-PT$^\dagger$~\cite{sam-pt}      & 83.1 & 83.2 & 82.9 & 77.6 & 74.8 & 80.4 & 74.0 & 73.3 & 76.0 & 70.0 & 76.7 & 4.1 & 2.1 & 6.0 \\
Matcher~\cite{matcher}      & 86.1 & 85.2 & \textbf{86.7} & 79.5 & 76.5 & 82.6 & - & - & - & - & - & - & - & - \\
\textbf{DRIFT(Ours)}       & \textbf{86.6} & \textbf{87.2} & 86.2 & \textbf{81.3} & \textbf{78.8} & \textbf{83.7} & \textbf{75.3} & 74.5 & 77.3 & \textbf{71.0} & \textbf{78.3} & \textbf{49.1} & \textbf{47.7} & \textbf{50.4} \\
\midrule
\multicolumn{15}{>{\columncolor{gray!20}}l}{\textit{Fully Supervised}} \\
\textcolor{gray}{CFBI~\cite{cfbi}}        & \textcolor{gray}{89.4} & \textcolor{gray}{88.3} & \textcolor{gray}{90.5} & \textcolor{gray}{81.9} & \textcolor{gray}{79.1} & \textcolor{gray}{84.6} & \textcolor{gray}{81.4} & \textcolor{gray}{81.1} & \textcolor{gray}{85.8} & \textcolor{gray}{75.3} & \textcolor{gray}{83.4} & \textcolor{gray}{53.5} & \textcolor{gray}{50.9} & \textcolor{gray}{56.1} \\
\textcolor{gray}{STCN~\cite{stcn}}        & \textcolor{gray}{91.6} & \textcolor{gray}{90.8} & \textcolor{gray}{92.5} & \textcolor{gray}{85.4} & \textcolor{gray}{82.2} & \textcolor{gray}{88.6} & \textcolor{gray}{83.0} & \textcolor{gray}{81.9} & \textcolor{gray}{86.5} & \textcolor{gray}{77.9} & \textcolor{gray}{85.7} & \textcolor{gray}{87.3} & \textcolor{gray}{85.4} & \textcolor{gray}{89.2} \\
\textcolor{gray}{AOT~\cite{aot}}         & \textcolor{gray}{91.1} & \textcolor{gray}{90.1} & \textcolor{gray}{92.1} & \textcolor{gray}{84.9} & \textcolor{gray}{82.3} & \textcolor{gray}{87.5} & \textcolor{gray}{85.5} & \textcolor{gray}{84.5} & \textcolor{gray}{89.5} & \textcolor{gray}{79.6} & \textcolor{gray}{88.2} & \textcolor{gray}{84.3} & \textcolor{gray}{83.2} & \textcolor{gray}{85.4} \\
\textcolor{gray}{XMem~\cite{xmem}}        & \textcolor{gray}{91.5} & \textcolor{gray}{90.4} & \textcolor{gray}{92.7} & \textcolor{gray}{86.2} & \textcolor{gray}{82.9} & \textcolor{gray}{89.5} & \textcolor{gray}{85.7} & \textcolor{gray}{84.6} & \textcolor{gray}{89.3} & \textcolor{gray}{80.2} & \textcolor{gray}{88.7} & \textcolor{gray}{89.8} & \textcolor{gray}{88.0} & \textcolor{gray}{91.6} \\
\textcolor{gray}{Cutie-base~\cite{cutie}}  & \textcolor{gray}{-}    & \textcolor{gray}{-}    & \textcolor{gray}{-}    & \textcolor{gray}{88.8} & \textcolor{gray}{85.4} & \textcolor{gray}{92.3} & \textcolor{gray}{86.1} & \textcolor{gray}{85.5} & \textcolor{gray}{90.0} & \textcolor{gray}{80.6} & \textcolor{gray}{88.3} & \textcolor{gray}{-}    & \textcolor{gray}{-}    & \textcolor{gray}{-}    \\
\bottomrule
\bottomrule
\end{tabular}
}
\label{tab:eval_vos}
\end{table*}

\noindent \textbf{Distribution of Learned Embeddings} \ \ 
\cref{fig:tsne} visualizes token embeddings of object class names and textual-inversion–learned embeddings aligned to specific objects.
We observe two clearly separated clusters, each grouping one type of embedding.
This indicates that the learned embeddings do not encode semantic information about the target object.
Instead, they behave as small, learnable parameters that allow fine-grained control over self-attention maps, effectively acting as tunable knobs for label propagation.

\noindent \textbf{Effect of Adaptive Head Weighting} \ \
We also assess the adaptive head weighting technique, comparing it to uniform averaging under two conditions: (i) with DDIM inversion and textual inversion (+DI+TI), and (ii) without them (Baseline).
The results show that learned weighting consistently achieves higher scores than uniform averaging across metrics. 
These improvements suggest that combining different heads attending to complementary regions yields more reliable mask propagation.

\subsection{Evaluation of \textbf{\drift{}}}
\label{sec:eval}

\begin{figure*}[t]
\centering
\includegraphics[width=\linewidth]{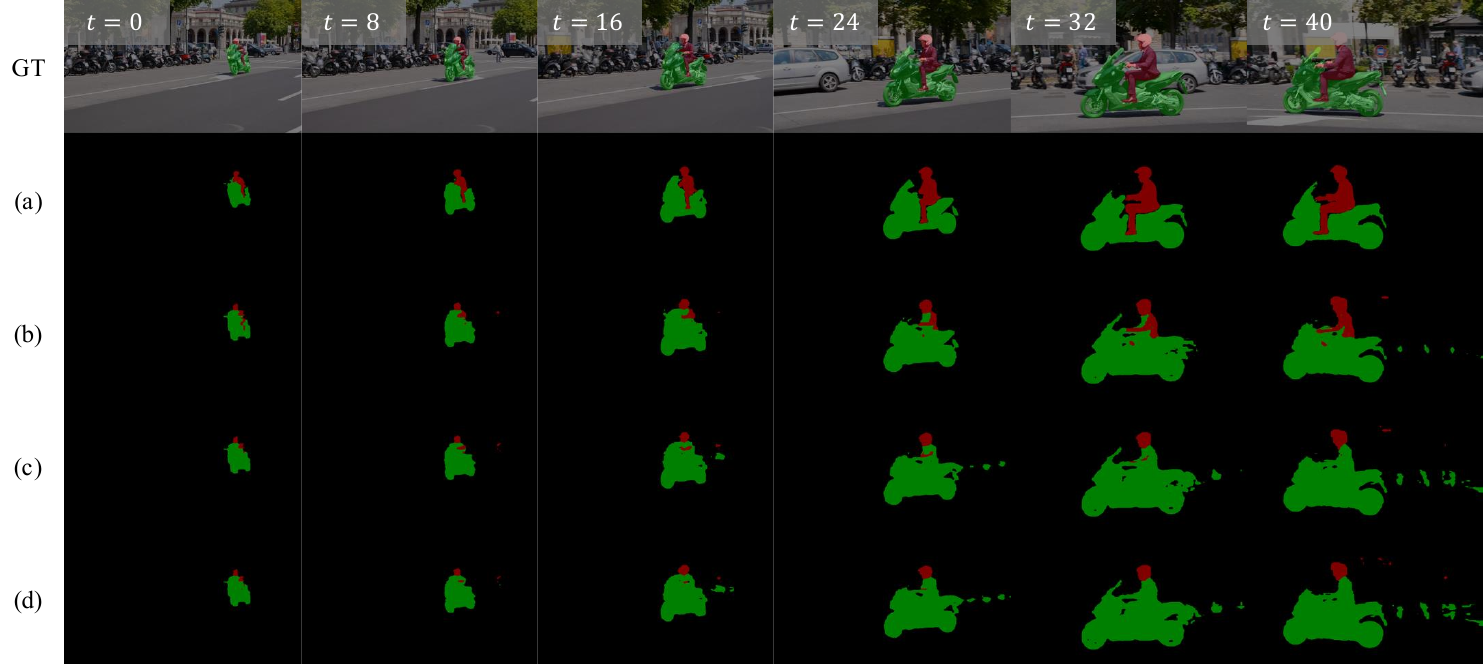}
\caption{%
\textbf{Qualitative Comparison of Model Variants on DAVIS 2017.} 
Each row shows segmentation results over time, with the top row showing GT masks and the others corresponding to the model ablations in \cref{tab:qual_ablation}.
Removing components leads to spatial drift or semantic ambiguity, while the full model (a) maintains accurate and coherent instance masks across frames.
}
\label{fig:qualitative}
\end{figure*}

\noindent \textbf{Comparisons to State-of-the-Art Methods} \ \ 
\cref{tab:eval_vos} presents a comparison between the proposed \drift{} and existing state-of-the-art methods. 
We evaluate two variants of \drift{}: a pure diffusion-based version without SAM~\citep{sam} and a SAM-integrated version. 
The former is compared against zero-shot baselines STC~\citep{stc}, DINO~\citep{dino}, and DIFT~\citep{dift}, 
while the latter is compared against SegGPT~\citep{seggpt}, SAM-PT~\citep{sam-pt}, and Matcher~\citep{matcher} which are zero-shot with image segmentation annotations.
Without SAM, our model outperforms three existing methods that also employ a label propagation approach but utilize different features. 
While DINO demonstrates remarkable performance without direct optimization on segmentation data by leveraging robust feature learning, our method shows even better results, achieving an average relative improvement of 5.94\% over DINO. 
When SAM is incorporated into our framework, the performance is further enhanced, achieving an average relative improvement of 5.59\% compared to ours without SAM.
Furthermore, our full \drift{} with SAM surpasses SegGPT—which is trained on large-scale image segmentation datasets—as well as SAM-PT and Matcher, both of which directly leverage SAM, achieving an average relative improvement of 3.6\%, 4.62\%, and 1.49\% across the short video benchmarks.
This highlights that diffusion-driven propagation outperforms SAM-PT and Matcher, despite both relying on SAM.
Both SAM-PT and SegGPT struggle to generalize to longer video sequences showing significant performance drops on Long Videos.
In contrast, \drift{} demonstrates its superior ability to maintain temporal coherence and instance identity across extended sequences, without requiring video-specific supervision.
Note that SAM-PT leverages additional supervision from a point tracking dataset~\citep{tap-vid}, which offers even denser annotations than labels for object tracking via segmentation. 
Despite this, \drift{} outperforms it across all benchmarks.
Finally, despite being a fully zero-shot approach, our method exhibits comparable scores to some fully-supervised methods, highlighting its remarkable generalization and effectiveness without relying on annotated training data.

\noindent \textbf{Ablation of Each Component} \ \ 
\cref{tab:qual_ablation} quantifies the impact of each component in \drift{}, with qualitative examples provided in \cref{fig:qualitative}.
Starting from the full model (a), which includes DDIM inversion (DI), textual inversion with adaptive head weighting (TI\&HW), and the SAM module (SAM), we observe the highest performance with 81.3\% in $\mathcal{J\&F}_\mathrm{m}$, reflecting precise boundary refinement and stable mask propagation. 
Removing each component leads to performance drops, illustrating the effectiveness of SAM for boundary refinement, textual inversion and head weighting for instance discrimination, and DDIM inversion for semantic stability.
The same trend is visible in the qualitative comparisons in \cref{fig:qualitative}, where each component contributes to more coherent and temporally consistent segmentation.
\begin{table}
    \centering
    \caption{\textbf{Ablation of \drift{} Components on DAVIS 2017.}
    Each row shows performance after removing one component from the configuration in the preceding row: DDIM inversion (DI), textual inversion (TI), head weighting (HW), and SAM refinement.
    }
    \scalebox{0.9}{
    \begin{tabular}{llccc}
        \toprule
        \toprule
         & Ablations & $\mathcal{J\&F}_\mathrm{m}$ & $\mathcal{J}_\mathrm{m}$ & $\mathcal{F}_\mathrm{m}$ \\
        \midrule
        (a) & \drift{} & 81.3 & 78.8 & 83.7 \\
        (b) & $-$SAM & 74.8 & 70.7 & 78.9 \\
        (c) & $-$TI\&HW & 71.8 & 67.9 & 75.6 \\
        (d) & $-$DI & 71.1 & 67.0 & 75.1 \\
        \bottomrule
        \bottomrule
    \end{tabular}
    }
    \vspace{-1em}
    \label{tab:qual_ablation}
\end{table}

\begin{figure}
\centering
    \begin{tikzpicture}
    \begin{groupplot}[
      group style={
        group size=1 by 2,
        vertical sep=0pt,
        xticklabels at=edge bottom,
        ylabels at=edge left,
      },
      width=\columnwidth,
      xmin=0, xmax=55,
      label style={font=\scriptsize},
      legend style={font=\scriptsize, draw=none, at={(0.5, 1.05)}, anchor=south, legend columns=-1},
    ]
    
    \nextgroupplot[
      height=2.8cm,
      ymin=74, ymax=82,
      ytick={78,82},
      tick label style={font=\scriptsize},
      axis x line=top,
      xtick align=inside,
      axis line style={-},       
      axis y discontinuity=parallel,
      clip marker paths=true,
      ymajorgrids,               
      grid style={gray!45},
      mark size=1.5pt
    ]
    
    \addplot+[mark=o, thick, restrict y to domain=-100:62, unbounded coords=discard]
    coordinates {(5,56.7) (10,59.3) (15,59.6) (20,60.4) (25,60.6)
                   (30,60.5) (35,60.7) (40,60.9) (45,61.1) (50,61.2)};
    \addlegendentry{$n$ = 1}
    \addplot+[mark=square, thick, restrict y to domain=74:100, unbounded coords=discard]
      coordinates {(5,78.75) (10,79.67) (15,80.2) (20,80.0) (25,80.2)
                   (30,80.33) (35,81.14) (40,81.34) (45,80.94) (50,80.64)};
    \addlegendentry{$n$ = 2}
    \addplot+[mark=diamond, thick, restrict y to domain=74:100, unbounded coords=discard]
      coordinates {(5,78.84) (10,79.72) (15,80.2) (20,79.95) (25,80.15)
                   (30,80.25) (35,81.05) (40,81.25) (45,80.85) (50,80.55)};
    \addlegendentry{$n$ = 3}
    
    \addplot+[mark=triangle, thick, restrict y to domain=74:100, unbounded coords=discard]
      coordinates {(5,78.0) (10,78.2) (15,78.2) (20,78.4) (25,79.0)
                   (30,78.5) (35,79.0) (40,78.4) (45,79.9) (50,79.3)};
    \addlegendentry{$n$ = 5}
    
    \nextgroupplot[
      xlabel={The number of candidates $(p)$},
      height=2.8cm,
      ymin=56, ymax=62,
      ytick={56,60},
      xtick={0,10,20,30,40,50},
      tick label style={font=\scriptsize},
      axis x line=bottom,
      xtick align=inside,
      axis line style={-},       
      clip marker paths=true,
      ymajorgrids,               
      grid style={gray!45},
      mark size=1.5pt
    ]
    
    \addplot+[mark=o, thick, restrict y to domain=-100:62, unbounded coords=discard]
      coordinates {(5,56.7) (10,59.3) (15,59.6) (20,60.4) (25,60.6)
                   (30,60.5) (35,60.7) (40,60.9) (45,61.1) (50,61.2)};
    \end{groupplot}
    \node[rotate=90, anchor=center, yshift=18pt, font=\scriptsize]
        at ($(group c1r1.west)!0.5!(group c1r2.west)$) {$\mathcal{J}\&\mathcal{F}_\mathrm{m}$};    
    \end{tikzpicture}
\captionsetup{aboveskip=-8pt, belowskip=0pt}
\caption{
    \textbf{Effect of the Number of Prompting Points $(n)$ for SAM and Candidate Masks $(p)$ on DAVIS 2017.}
    Performance improves with larger $p$, peaking at $p{=}40$, and saturates around $n{=}3$, with a slight drop at $n{=}5$.}
\label{fig:abl_sam}
\end{figure}
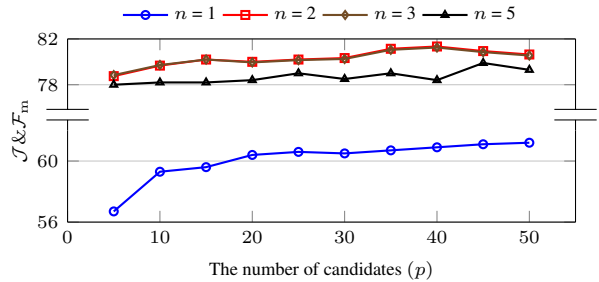

\noindent \textbf{Point Sampling for Prompting SAM} \ \ 
Finally, we investigate how the contribution of SAM~\citep{sam} to the segmentation quality varies with the number of sampled points ($n$) and mask candidates ($p$), as shown in \cref{fig:abl_sam}.
When $n{=}1$, performance is poor because a single prompt often produces unstable segments, covering only part of the object or an overly large area.
With $n{=}2$, stability improves substantially, leading to a significant performance gain, which then saturates at $n{=}3$.
Adding more points (\eg, $n{=}5$) degrades performance, as the increased number of prompts raises the likelihood of including mislabeled regions.
Increasing $p$ also improves performance, but the gains saturate at around $p{=}40$.
Overall, with $n{=}2$ points and $p{=}40$ candidates, we achieve the best performance of 81.3\% in $\mathcal{J}\&\mathcal{F}_\mathrm{m}$, with a improvement over 74.8\% without refinement.

\section{Conclusion}
We show that image diffusion models, without video supervision, naturally support temporal reasoning when self-attention is repurposed as a cross-frame propagation kernel. 
We demonstrate that test-time optimizations—DDIM inversion, mask-specific textual inversion, and adaptive head weighting—lead to more reliable cross-frame label propagation.
Building on these insights, we introduced \drift{}, a framework for object tracking in videos using a pretrained image diffusion model with SAM-guided mask refinement. 
\drift{} achieves state-of-the-art zero-shot performance on standard video object segmentation benchmarks.
These results highlight the potential of diffusion models as a foundation for broader video understanding tasks.
{
    \small
    \bibliographystyle{ieeenat_fullname}
    \bibliography{main}

@String(CVPR= {IEEE Conf. Comput. Vis. Pattern Recog.})

@String(ICCV= {Int. Conf. Comput. Vis.})

@String(CVPR  = {CVPR})

@String(ICCV  = {ICCV})

@InProceedings{stablediffusion,
    author    = {Rombach, Robin and Blattmann, Andreas and Lorenz, Dominik and Esser, Patrick and Ommer, Bj\"orn},
    title     = {High-Resolution Image Synthesis With Latent Diffusion Models},
    booktitle = {Proceedings of the IEEE/CVF Conference on Computer Vision and Pattern Recognition (CVPR)},
    month     = {June},
    year      = {2022},
    pages     = {10684-10695}
}

@InProceedings{sam,
    author    = {Kirillov, Alexander and Mintun, Eric and Ravi, Nikhila and Mao, Hanzi and Rolland, Chloe and Gustafson, Laura and Xiao, Tete and Whitehead, Spencer and Berg, Alexander C. and Lo, Wan-Yen and Dollar, Piotr and Girshick, Ross},
    title     = {Segment Anything},
    booktitle = {Proceedings of the IEEE/CVF International Conference on Computer Vision (ICCV)},
    month     = {October},
    year      = {2023},
    pages     = {4015-4026}
}

@misc{textual_inversion,
      doi = {10.48550/ARXIV.2208.01618},
      url = {https://arxiv.org/abs/2208.01618},
      author = {Gal, Rinon and Alaluf, Yuval and Atzmon, Yuval and Patashnik, Or and Bermano, Amit H. and Chechik, Gal and Cohen-Or, Daniel},
      title = {An Image is Worth One Word: Personalizing Text-to-Image Generation using Textual Inversion},
      publisher = {arXiv},
      year = {2022},
      primaryClass={cs.CV}
}

@misc{diffsegmenter,
      title={Diffusion Model is Secretly a Training-free Open Vocabulary Semantic Segmenter}, 
      author={Jinglong Wang and Xiawei Li and Jing Zhang and Qingyuan Xu and Qin Zhou and Qian Yu and Lu Sheng and Dong Xu},
      year={2024},
      eprint={2309.02773},
      archivePrefix={arXiv},
      primaryClass={cs.CV},
      url={https://arxiv.org/abs/2309.02773}, 
}

@InProceedings{diffseg,
    author    = {Tian, Junjiao and Aggarwal, Lavisha and Colaco, Andrea and Kira, Zsolt and Gonzalez-Franco, Mar},
    title     = {Diffuse Attend and Segment: Unsupervised Zero-Shot Segmentation using Stable Diffusion},
    booktitle = {Proceedings of the IEEE/CVF Conference on Computer Vision and Pattern Recognition (CVPR)},
    month     = {June},
    year      = {2024},
    pages     = {3554-3563}
}

@article{ddim,
  title={Denoising Diffusion Implicit Models},
  author={Song, Jiaming and Meng, Chenlin and Ermon, Stefano},
  journal={arXiv:2010.02502},
  year={2020},
  month={October},
  abbr={Preprint},
  url={https://arxiv.org/abs/2010.02502}
}

@inproceedings{diffusionbeatgan,
 author = {Dhariwal, Prafulla and Nichol, Alexander},
 booktitle = {Advances in Neural Information Processing Systems},
 editor = {M. Ranzato and A. Beygelzimer and Y. Dauphin and P.S. Liang and J. Wortman Vaughan},
 pages = {8780--8794},
 publisher = {Curran Associates, Inc.},
 title = {Diffusion Models Beat GANs on Image Synthesis},
 url = {https://proceedings.neurips.cc/paper_files/paper/2021/file/49ad23d1ec9fa4bd8d77d02681df5cfa-Paper.pdf},
 volume = {34},
 year = {2021}
}

@inproceedings{diffcut,
 author = {Couairon, Paul and Shukor, Mustafa and Haugeard, Jean-Emmanuel and Cord, Matthieu and Thome, Nicolas},
 booktitle = {Advances in Neural Information Processing Systems},
 editor = {A. Globerson and L. Mackey and D. Belgrave and A. Fan and U. Paquet and J. Tomczak and C. Zhang},
 pages = {13548--13578},
 publisher = {Curran Associates, Inc.},
 title = {DiffCut: Catalyzing Zero-Shot Semantic Segmentation with Diffusion Features and Recursive Normalized Cut},
 url = {https://proceedings.neurips.cc/paper_files/paper/2024/file/1867748a011e1425b924ec72a4066b62-Paper-Conference.pdf},
 volume = {37},
 year = {2024}
}

@inproceedings{stc,
 author = {Jabri, Allan and Owens, Andrew and Efros, Alexei},
 booktitle = {Advances in Neural Information Processing Systems},
 editor = {H. Larochelle and M. Ranzato and R. Hadsell and M.F. Balcan and H. Lin},
 pages = {19545--19560},
 publisher = {Curran Associates, Inc.},
 title = {Space-Time Correspondence as a Contrastive Random Walk},
 url = {https://proceedings.neurips.cc/paper_files/paper/2020/file/e2ef524fbf3d9fe611d5a8e90fefdc9c-Paper.pdf},
 volume = {33},
 year = {2020}
}

@InProceedings{dino,
    author    = {Caron, Mathilde and Touvron, Hugo and Misra, Ishan and J\'egou, Herv\'e and Mairal, Julien and Bojanowski, Piotr and Joulin, Armand},
    title     = {Emerging Properties in Self-Supervised Vision Transformers},
    booktitle = {Proceedings of the IEEE/CVF International Conference on Computer Vision (ICCV)},
    month     = {October},
    year      = {2021},
    pages     = {9650-9660}
}

@InProceedings{davis2016,
author = {Perazzi, Federico and Pont-Tuset, Jordi and McWilliams, Brian and Van Gool, Luc and Gross, Markus and Sorkine-Hornung, Alexander},
title = {A Benchmark Dataset and Evaluation Methodology for Video Object Segmentation},
booktitle = {Proceedings of the IEEE Conference on Computer Vision and Pattern Recognition (CVPR)},
month = {June},
year = {2016}
}

@article{davis2017,
  author = {Jordi Pont-Tuset and Federico Perazzi and Sergi Caelles and Pablo Arbel\'aez and Alexander Sorkine-Hornung and Luc {Van Gool}},
  title = {The 2017 DAVIS Challenge on Video Object Segmentation},
  journal = {arXiv:1704.00675},
  year = {2017}
}

@inproceedings{longvideos,
 author = {Liang, Yongqing and Li, Xin and Jafari, Navid and Chen, Jim},
 booktitle = {Advances in Neural Information Processing Systems},
 editor = {H. Larochelle and M. Ranzato and R. Hadsell and M.F. Balcan and H. Lin},
 pages = {3430--3441},
 publisher = {Curran Associates, Inc.},
 title = {Video Object Segmentation with Adaptive Feature Bank and Uncertain-Region Refinement},
 url = {https://proceedings.neurips.cc/paper/2020/file/234833147b97bb6aed53a8f4f1c7a7d8-Paper.pdf},
 volume = {33},
 year = {2020}
}

@InProceedings{seggpt,
    author    = {Wang, Xinlong and Zhang, Xiaosong and Cao, Yue and Wang, Wen and Shen, Chunhua and Huang, Tiejun},
    title     = {SegGPT: Towards Segmenting Everything in Context},
    booktitle = {Proceedings of the IEEE/CVF International Conference on Computer Vision (ICCV)},
    month     = {October},
    year      = {2023},
    pages     = {1130-1140}
}

@InProceedings{sam-pt,
    author    = {Raji\v{c}, Frano and Ke, Lei and Tai, Yu-Wing and Tang, Chi-Keung and Danelljan, Martin and Yu, Fisher},
    title     = {Segment Anything Meets Point Tracking},
    booktitle = {Proceedings of the Winter Conference on Applications of Computer Vision (WACV)},
    month     = {February},
    year      = {2025},
    pages     = {9284-9293}
}

@inproceedings{cfbi,
  title={Collaborative video object segmentation by foreground-background integration},
  author={Yang, Zongxin and Wei, Yunchao and Yang, Yi},
  booktitle={European Conference on Computer Vision},
  pages={332--348},
  year={2020},
  organization={Springer}
}

@inproceedings{stcn,
 author = {Cheng, Ho Kei and Tai, Yu-Wing and Tang, Chi-Keung},
 booktitle = {Advances in Neural Information Processing Systems},
 editor = {M. Ranzato and A. Beygelzimer and Y. Dauphin and P.S. Liang and J. Wortman Vaughan},
 pages = {11781--11794},
 publisher = {Curran Associates, Inc.},
 title = {Rethinking Space-Time Networks with Improved Memory Coverage for Efficient Video Object Segmentation},
 url = {https://proceedings.neurips.cc/paper_files/paper/2021/file/61b4a64be663682e8cb037d9719ad8cd-Paper.pdf},
 volume = {34},
 year = {2021}
}

@inproceedings{aot,
 author = {Yang, Zongxin and Wei, Yunchao and Yang, Yi},
 booktitle = {Advances in Neural Information Processing Systems},
 editor = {M. Ranzato and A. Beygelzimer and Y. Dauphin and P.S. Liang and J. Wortman Vaughan},
 pages = {2491--2502},
 publisher = {Curran Associates, Inc.},
 title = {Associating Objects with Transformers for Video Object Segmentation},
 url = {https://proceedings.neurips.cc/paper_files/paper/2021/file/147702db07145348245dc5a2f2fe5683-Paper.pdf},
 volume = {34},
 year = {2021}
}

@inproceedings{xmem,
  title={Xmem: Long-term video object segmentation with an atkinson-shiffrin memory model},
  author={Cheng, Ho Kei and Schwing, Alexander G},
  booktitle={European Conference on Computer Vision},
  pages={640--658},
  year={2022},
  organization={Springer}
}

@inproceedings{cutie,
  title={Putting the object back into video object segmentation},
  author={Cheng, Ho Kei and Oh, Seoung Wug and Price, Brian and Lee, Joon-Young and Schwing, Alexander},
  booktitle={Proceedings of the IEEE/CVF Conference on Computer Vision and Pattern Recognition},
  pages={3151--3161},
  year={2024}
}

@inproceedings{cotracker,
  title={Cotracker: It is better to track together},
  author={Karaev, Nikita and Rocco, Ignacio and Graham, Benjamin and Neverova, Natalia and Vedaldi, Andrea and Rupprecht, Christian},
  booktitle={European Conference on Computer Vision},
  pages={18--35},
  year={2024},
  organization={Springer}
}

@article{ytvos2018,
  author       = {Ning Xu and
                  Linjie Yang and
                  Yuchen Fan and
                  Dingcheng Yue and
                  Yuchen Liang and
                  Jianchao Yang and
                  Thomas S. Huang},
  title        = {YouTube-VOS: {A} Large-Scale Video Object Segmentation Benchmark},
  journal      = {CoRR},
  volume       = {abs/1809.03327},
  year         = {2018},
  url          = {http://arxiv.org/abs/1809.03327},
}

@article{imagen,
  title={Photorealistic text-to-image diffusion models with deep language understanding},
  author={Saharia, Chitwan and Chan, William and Saxena, Saurabh and Li, Lala and Whang, Jay and Denton, Emily L and Ghasemipour, Kamyar and Gontijo Lopes, Raphael and Karagol Ayan, Burcu and Salimans, Tim and others},
  journal={Advances in neural information processing systems},
  volume={35},
  pages={36479--36494},
  year={2022}
}

@article{tap-vid,
  title={Tap-vid: A benchmark for tracking any point in a video},
  author={Doersch, Carl and Gupta, Ankush and Markeeva, Larisa and Recasens, Adria and Smaira, Lucas and Aytar, Yusuf and Carreira, Joao and Zisserman, Andrew and Yang, Yi},
  journal={Advances in Neural Information Processing Systems},
  volume={35},
  pages={13610--13626},
  year={2022}
}

@article{dalle2,
  title={Hierarchical text-conditional image generation with clip latents},
  author={Ramesh, Aditya and Dhariwal, Prafulla and Nichol, Alex and Chu, Casey and Chen, Mark},
  journal={arXiv preprint arXiv:2204.06125},
  volume={1},
  number={2},
  pages={3},
  year={2022}
}

@inproceedings{vd-it,
  title={Exploring pre-trained text-to-video diffusion models for referring video object segmentation},
  author={Zhu, Zixin and Feng, Xuelu and Chen, Dongdong and Yuan, Junsong and Qiao, Chunming and Hua, Gang},
  booktitle={European Conference on Computer Vision},
  pages={452--469},
  year={2024},
  organization={Springer}
}

@misc{vidseg,
      title={Zero-Shot Video Semantic Segmentation based on Pre-Trained Diffusion Models}, 
      author={Qian Wang and Abdelrahman Eldesokey and Mohit Mendiratta and Fangneng Zhan and Adam Kortylewski and Christian Theobalt and Peter Wonka},
      year={2024},
      eprint={2405.16947},
      archivePrefix={arXiv},
      primaryClass={cs.CV}
}

@article{dift,
  title={Emergent correspondence from image diffusion},
  author={Tang, Luming and Jia, Menglin and Wang, Qianqian and Phoo, Cheng Perng and Hariharan, Bharath},
  journal={Advances in Neural Information Processing Systems},
  volume={36},
  pages={1363--1389},
  year={2023}
}

@inproceedings{paccrf,
  title={Pixel-adaptive convolutional neural networks},
  author={Su, Hang and Jampani, Varun and Sun, Deqing and Gallo, Orazio and Learned-Miller, Erik and Kautz, Jan},
  booktitle={Proceedings of the IEEE/CVF Conference on Computer Vision and Pattern Recognition},
  pages={11166--11175},
  year={2019}
}

@article{uziel2023vit,
  title={From vit features to training-free video object segmentation via streaming-data mixture models},
  author={Uziel, Roy and Dinari, Or and Freifeld, Oren},
  journal={Advances in Neural Information Processing Systems},
  volume={36},
  pages={10995--11007},
  year={2023}
}

@article{ddpm,
  title={Denoising diffusion probabilistic models},
  author={Ho, Jonathan and Jain, Ajay and Abbeel, Pieter},
  journal={Advances in neural information processing systems},
  volume={33},
  pages={6840--6851},
  year={2020}
}

@inproceedings{dpm,
  title={Deep unsupervised learning using nonequilibrium thermodynamics},
  author={Sohl-Dickstein, Jascha and Weiss, Eric and Maheswaranathan, Niru and Ganguli, Surya},
  booktitle={International conference on machine learning},
  pages={2256--2265},
  year={2015},
  organization={pmlr}
}

@article{sdxl,
  title={Sdxl: Improving latent diffusion models for high-resolution image synthesis},
  author={Podell, Dustin and English, Zion and Lacey, Kyle and Blattmann, Andreas and Dockhorn, Tim and M{\"u}ller, Jonas and Penna, Joe and Rombach, Robin},
  journal={arXiv preprint arXiv:2307.01952},
  year={2023}
}

@inproceedings{blip2,
  title={Blip-2: Bootstrapping language-image pre-training with frozen image encoders and large language models},
  author={Li, Junnan and Li, Dongxu and Savarese, Silvio and Hoi, Steven},
  booktitle={International conference on machine learning},
  pages={19730--19742},
  year={2023},
  organization={PMLR}
}

@article{flatten,
  title={FLATTEN: optical FLow-guided ATTENtion for consistent text-to-video editing},
  author={Cong, Yuren and Xu, Mengmeng and Simon, Christian and Chen, Shoufa and Ren, Jiawei and Xie, Yanping and Perez-Rua, Juan-Manuel and Rosenhahn, Bodo and Xiang, Tao and He, Sen},
  journal={arXiv preprint arXiv:2310.05922},
  year={2023}
}

@inproceedings{difftracker,
  title={Diff-tracker: text-to-image diffusion models are unsupervised trackers},
  author={Zhang, Zhengbo and Xu, Li and Peng, Duo and Rahmani, Hossein and Liu, Jun},
  booktitle={European Conference on Computer Vision},
  pages={319--337},
  year={2024},
  organization={Springer}
}

@inproceedings{
smite,
title={{SMITE}: Segment Me In TimE},
author={Amirhossein Alimohammadi and Sauradip Nag and Saeid Asgari and Andrea Tagliasacchi and Ghassan Hamarneh and Ali Mahdavi Amiri},
booktitle={The Thirteenth International Conference on Learning Representations},
year={2025},
}

@inproceedings{
hedlin2023unsupervised,
title={Unsupervised Semantic Correspondence Using Stable Diffusion},
author={Eric Hedlin and Gopal Sharma and Shweta Mahajan and Hossam Isack and Abhishek Kar and Andrea Tagliasacchi and Kwang Moo Yi},
booktitle={Thirty-seventh Conference on Neural Information Processing Systems},
year={2023},
}

@inproceedings{odise,
  title={Open-vocabulary panoptic segmentation with text-to-image diffusion models},
  author={Xu, Jiarui and Liu, Sifei and Vahdat, Arash and Byeon, Wonmin and Wang, Xiaolong and De Mello, Shalini},
  booktitle={Proceedings of the IEEE/CVF conference on computer vision and pattern recognition},
  pages={2955--2966},
  year={2023}
}

@article{diffusionhyperfeatures,
  title={Diffusion hyperfeatures: Searching through time and space for semantic correspondence},
  author={Luo, Grace and Dunlap, Lisa and Park, Dong Huk and Holynski, Aleksander and Darrell, Trevor},
  journal={Advances in Neural Information Processing Systems},
  volume={36},
  pages={47500--47510},
  year={2023}
}

@inproceedings{generalized_diffdet,
  title={Generalized Diffusion Detector: Mining Robust Features from Diffusion Models for Domain-Generalized Detection},
  author={He, Boyong and Ji, Yuxiang and Ye, Qianwen and Tan, Zhuoyue and Wu, Liaoni},
  booktitle={Proceedings of the Computer Vision and Pattern Recognition Conference},
  pages={9921--9932},
  year={2025}
}

@inproceedings{wang2019learning,
  title={Learning correspondence from the cycle-consistency of time},
  author={Wang, Xiaolong and Jabri, Allan and Efros, Alexei A},
  booktitle={Proceedings of the IEEE/CVF conference on computer vision and pattern recognition},
  pages={2566--2576},
  year={2019}
}

@article{lai2019self,
  title={Self-supervised learning for video correspondence flow},
  author={Lai, Zihang and Xie, Weidi},
  journal={arXiv preprint arXiv:1905.00875},
  year={2019}
}

@inproceedings{li2016unsupervised,
  title={Unsupervised learning of edges},
  author={Li, Yin and Paluri, Manohar and Rehg, James M and Doll{\'a}r, Piotr},
  booktitle={Proceedings of the IEEE conference on computer vision and pattern recognition},
  pages={1619--1627},
  year={2016}
}

@inproceedings{qi2023fatezero,
  title={Fatezero: Fusing attentions for zero-shot text-based video editing},
  author={Qi, Chenyang and Cun, Xiaodong and Zhang, Yong and Lei, Chenyang and Wang, Xintao and Shan, Ying and Chen, Qifeng},
  booktitle={Proceedings of the IEEE/CVF International Conference on Computer Vision},
  pages={15932--15942},
  year={2023}
}

@inproceedings{feng2024wave,
  title={Wave: Warping ddim inversion features for zero-shot text-to-video editing},
  author={Feng, Yutang and Gao, Sicheng and Bao, Yuxiang and Wang, Xiaodi and Han, Shumin and Zhang, Juan and Zhang, Baochang and Yao, Angela},
  booktitle={European Conference on Computer Vision},
  pages={38--55},
  year={2024},
  organization={Springer}
}

@article{song2020score,
  title={Score-based generative modeling through stochastic differential equations},
  author={Song, Yang and Sohl-Dickstein, Jascha and Kingma, Diederik P and Kumar, Abhishek and Ermon, Stefano and Poole, Ben},
  journal={arXiv preprint arXiv:2011.13456},
  year={2020}
}

@inproceedings{dit,
  title={Scalable diffusion models with transformers},
  author={Peebles, William and Xie, Saining},
  booktitle={Proceedings of the IEEE/CVF international conference on computer vision},
  pages={4195--4205},
  year={2023}
}

@inproceedings{dreambooth,
  title={Dreambooth: Fine tuning text-to-image diffusion models for subject-driven generation},
  author={Ruiz, Nataniel and Li, Yuanzhen and Jampani, Varun and Pritch, Yael and Rubinstein, Michael and Aberman, Kfir},
  booktitle={Proceedings of the IEEE/CVF conference on computer vision and pattern recognition},
  pages={22500--22510},
  year={2023}
}

@inproceedings{videoldm,
  title={Align your latents: High-resolution video synthesis with latent diffusion models},
  author={Blattmann, Andreas and Rombach, Robin and Ling, Huan and Dockhorn, Tim and Kim, Seung Wook and Fidler, Sanja and Kreis, Karsten},
  booktitle={Proceedings of the IEEE/CVF conference on computer vision and pattern recognition},
  pages={22563--22575},
  year={2023}
}

@article{videocrafter,
  title={Videocrafter1: Open diffusion models for high-quality video generation},
  author={Chen, Haoxin and Xia, Menghan and He, Yingqing and Zhang, Yong and Cun, Xiaodong and Yang, Shaoshu and Xing, Jinbo and Liu, Yaofang and Chen, Qifeng and Wang, Xintao and others},
  journal={arXiv preprint arXiv:2310.19512},
  year={2023}
}

@article{animatediff,
  title={Animatediff: Animate your personalized text-to-image diffusion models without specific tuning},
  author={Guo, Yuwei and Yang, Ceyuan and Rao, Anyi and Liang, Zhengyang and Wang, Yaohui and Qiao, Yu and Agrawala, Maneesh and Lin, Dahua and Dai, Bo},
  journal={arXiv preprint arXiv:2307.04725},
  year={2023}
}

@article{matcher,
  title={Matcher: Segment anything with one shot using all-purpose feature matching},
  author={Liu, Yang and Zhu, Muzhi and Li, Hengtao and Chen, Hao and Wang, Xinlong and Shen, Chunhua},
  journal={arXiv preprint arXiv:2305.13310},
  year={2023}
}

@article{diffews,
  title={Unleashing the potential of the diffusion model in few-shot semantic segmentation},
  author={Zhu, Muzhi and Liu, Yang and Luo, Zekai and Jing, Chenchen and Chen, Hao and Xu, Guangkai and Wang, Xinlong and Shen, Chunhua},
  journal={Advances in Neural Information Processing Systems},
  volume={37},
  pages={42672--42695},
  year={2024}
}

@inproceedings{wu2023tune,
  title={Tune-a-video: One-shot tuning of image diffusion models for text-to-video generation},
  author={Wu, Jay Zhangjie and Ge, Yixiao and Wang, Xintao and Lei, Stan Weixian and Gu, Yuchao and Shi, Yufei and Hsu, Wynne and Shan, Ying and Qie, Xiaohu and Shou, Mike Zheng},
  booktitle={Proceedings of the IEEE/CVF international conference on computer vision},
  pages={7623--7633},
  year={2023}
}

@article{stablevideodiffusion,
  title={Stable video diffusion: Scaling latent video diffusion models to large datasets},
  author={Blattmann, Andreas and Dockhorn, Tim and Kulal, Sumith and Mendelevitch, Daniel and Kilian, Maciej and Lorenz, Dominik and Levi, Yam and English, Zion and Voleti, Vikram and Letts, Adam and others},
  journal={arXiv preprint arXiv:2311.15127},
  year={2023}
}

@article{nam2025emergent,
  title={Emergent Temporal Correspondences from Video Diffusion Transformers},
  author={Nam, Jisu and Son, Soowon and Chung, Dahyun and Kim, Jiyoung and Jin, Siyoon and Hur, Junhwa and Kim, Seungryong},
  journal={arXiv preprint arXiv:2506.17220},
  year={2025}
}
}
\clearpage
\maketitlesupplementary
\appendix

\begin{figure*}[!t]
    \centering
    \includegraphics[width=0.95\linewidth]{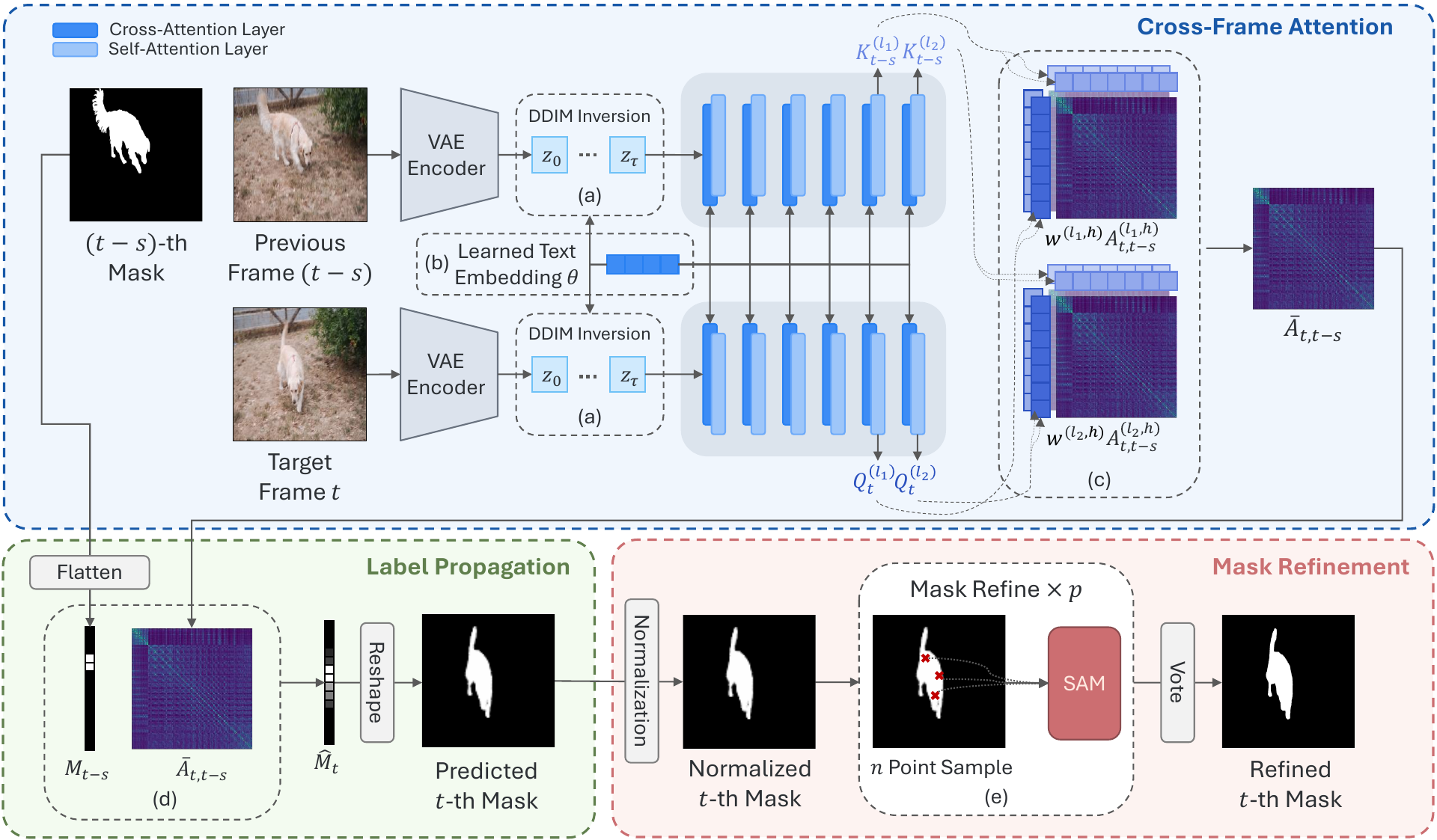}
    \caption{
    \textbf{Overall Pipeline of \drift{}.}
    Given a target frame $t$ and a reference frame $t{-}s$ with its mask, 
    (a) we perform DDIM inversion to obtain semantic latent representations $\mathbf{z}_\tau$.
    (b) Text embeddings $\theta$ are learned for the mask and fed to the diffusion model. 
    Cross-frame attention maps are then computed by matching queries from the target frame with keys from the reference frame across selected layers,
    and (c) the multi-head attention maps are aggregated using head-specific weights.
    (d) The aggregated cross-frame attention map $\bar{A}_{t,t-s}$ is used to propagate the mask from frame $t{-}s$ to $t$.
    (e) Finally, the obtained soft mask is refined using SAM resulting in a fine-grained mask in the target frame.
    }
    \label{fig:arch}
\end{figure*}
\section{Additional Details}

\subsection{Implementation Details}
\label{appendix:details}

We adopt Stable Diffusion 2.1~\citep{stablediffusion} as our backbone, which takes $768 \times 768$ images and produces $96 \times 96$ latent features. 
Self-attention maps are extracted from the first attention layer in the final decoder block (\path{up_blocks.3.attentions.0.transformer_blocks.0.attn1}).
At test time, we jointly optimize a mask-specific text token $\theta$ and head weights $w^{(l,h)}$ for each instance using Adam (lr $1\mathrm{e}{-4}$, 3{,}500 steps). 

Each instance introduces only $78{,}848$ learnable parameters for the text token and $5$ parameters for head weights, totaling $78{,}853$ parameters. 
DDIM inversion targets timestep $\tau = 41$ using 50 steps from the 1000-step schedule.
For refinement, we use SAM~\citep{sam} (ViT-H) with $n = 2$ point prompts sampled from the normalized soft mask, generating $p = 40$ candidate masks; the one with highest IoU is selected.
Label propagation uses the 7 most recent frames with initial frame. We apply spatial masking with radius $r = 14$ and retain the top $k = 15$ attention scores per query.
All experiments are conducted on NVIDIA H100 GPUs.
We also provide video qualitative results in the supplementary material.

\subsection{Evaluation Dataset Details}
\label{appendix:data}
\paragraph{DAVIS-2016~\cite{davis2016}.}  
The DAVIS-2016 dataset was originally introduced for single-object video object segmentation.  
The validation set consists of 20 high-quality video sequences (480p resolution), each annotated with a single foreground object mask at every frame.  
On average, each sequence contains around 70 frames, and the dataset is widely used to benchmark single-object VOS methods due to its precise, per-frame annotations.

\paragraph{DAVIS-2017~\cite{davis2017}.}  
DAVIS-2017 extends DAVIS-2016 by introducing multiple annotated objects per video, thereby increasing the difficulty of the segmentation task.  
The validation set contains 30 sequences with a total of 59 annotated objects, with around 70 frames per video.  
All frames are annotated, and the dataset is widely considered the standard benchmark for multi-object semi-supervised VOS evaluation.

\paragraph{YouTube-VOS 2018~\cite{ytvos2018}.}  
YouTube-VOS is the largest public benchmark for video object segmentation.  
The 2018 validation set includes 474 video sequences, covering 91 object categories, of which 65 are seen during training and 26 are unseen.  
Unlike DAVIS, annotations are provided every 5th frame instead of every frame.  
This yields a total of 12,593 annotated frames in the validation split.  
The large scale and category diversity of YouTube-VOS make it a challenging and comprehensive benchmark, particularly for evaluating generalization to unseen categories.

\paragraph{Long Videos~\cite{longvideos}.}  
The Long Videos dataset was designed to test the robustness of segmentation methods on long-duration sequences.  
It contains 3 validation videos, each with an average length of about 2,470 frames, far exceeding the sequence lengths of DAVIS or YouTube-VOS.  
For evaluation, 20 frames are uniformly sampled from each video and manually annotated with object masks.  
This setup allows the benchmark to focus on assessing temporal consistency and robustness of segmentation methods under extended time horizons.

\subsection{Soft IoU Metric}
\label{appendix:iou}
To select the most accurate refinement from multiple mask candidates generated by SAM~\cite{sam}, we measure the similarity between each binary SAM mask and the original soft mask prediction using a soft IoU metric. Given a soft mask $A \in [0, 1]^{H \times W}$ and a binary candidate mask $B \in \{0, 1\}^{H \times W}$, the soft IoU is computed as:
\begin{align}
    \mathrm{IoU}(A, B) = \frac{\sum_{i,j} \min(A_{i,j}, B_{i,j})}{\sum_{i,j} \max(A_{i,j}, B_{i,j}) + \epsilon},
\end{align}
where $\epsilon$ is a small constant added to the denominator for numerical stability.

Unlike the standard IoU computation that uses discrete set operations, soft masks represent confidence values or probability distributions over space. Therefore, we interpret the intersection and union between soft and binary masks as element-wise $\min$ and $\max$ operations, respectively. This formulation retains the probabilistic nature of the soft mask while enabling consistent comparison with discrete predictions.
Before computing soft IoU, we normalize the soft mask $A$ such that its values sum to one across spatial dimensions, treating it as a spatial probability distribution. To generate $p$ candidate masks, we sample $n$ point prompts from the normalized soft mask $p$ times, each used as a prompt to SAM. The mask with the highest soft IoU score against the original soft prediction is selected as the final output.

\subsection{Details of BLIP-2 Captioning}
\label{appendix:blip2}
We use BLIP-2 \citep{blip2} to generate noun-phrase captions for annotated instances in the first frame. Each object is cropped from the DAVIS mask with a small margin, and the background is masked out so that only the target region remains visible. The masked crops are then passed to BLIP-2 with the prompt, \texttt{"Question: Provide a short noun phrase that names only the main object in the image.\textbackslash n Answer:"}. The resulting captions are assigned to the corresponding instances and stored as their initial prompts. 

\section{Further Analayses}
\subsection{Limitation of Cosine Similarity under Textual Inversion}
\label{appendix:cos_ti}
As illustrated in \cref{fig:cos_vs_attn}, cosine similarity maps derived from raw diffusion features are noisy, often highlighting irrelevant regions in addition to the target object. 
Ideally, an affinity measure should emphasize the target region while suppressing unrelated areas, but cosine similarity lacks this selectivity. 
Moreover, in our setup only the textual tokens are optimized during textual inversion, while the diffusion backbone remains frozen. 
This limited degree of freedom makes it difficult to correct the noisy propagation induced by cosine similarity, highlighting the advantage of using self-attention maps that already encode semantically structured affinities.

\subsection{Component-wise Runtime Analysis}
\label{appendix:runtime}
\begin{table*}[t]
\centering
\caption{\textbf{Comparison of Execution Time and Performance across Different Models and Variants on DAVIS 2017.}
Runtime and accuracy comparison on the DAVIS~\citep{davis2017} validation set.
Execution time is reported per object on a single NVIDIA H100 GPU.
\drift{} is decomposed into cumulative components, where each row adds one module.
}
\scalebox{0.9}{
\begin{tabular}{cllccc}
\toprule
\toprule
 & Models & Execution Time (s/object) & $\mathcal{J\&F}_\mathrm{m}$ & $\mathcal{J}_\mathrm{m}$ & $\mathcal{F}_\mathrm{m}$\\
\midrule
1 & DINO (ViT-B/8) & 11.410 & 71.4 & 67.9 & 74.9\\
2 & SAM-PT (CoTracker + SAM-H) & 13.744 & 77.6 & 74.8 & 80.4\\
\midrule
3 & DRIFT (cross-frame propagation only) & 9.853 & 71.1 & 67.0 & 75.1\\
4 & \quad + DDIM inversion & \quad + 2.185 & 71.8 & 67.9 & 75.6\\
5 & \quad + SAM & \quad + 5.776 & 80.7 & 78.2 & 83.2\\
6 & \quad + Textual inversion \& Head weighting & \quad + 141.437 & 81.3 & 78.8 & 83.7\\
\bottomrule
\bottomrule
\end{tabular}
}
\label{tab:runtime}
\end{table*}
We report the average processing time per object on the DAVIS~\citep{davis2017} validation set, measured on a single NVIDIA H100 GPU. \cref{tab:runtime} presents both accuracy and runtime for DRIFT and relevant baselines. Our results are shown in a component-wise manner, where the total runtime corresponds to the sum of the relevant modules. The base propagation step in \drift{} achieves higher accuracy than existing trackers such as DINO (rows 1 and 4) and SAM-PT (rows 2 and 5), while maintaining comparable runtime efficiency. Incorporating DDIM inversion and SAM refinement leads to steady accuracy gains, with runtime overheads of 0.6s and 4.1s per object, respectively. Textual inversion adds a larger overhead, but it is performed only once on the first frame, making its relative cost less dominant for longer sequences. Importantly, even without textual inversion or SAM refinement, \drift{} achieves stronger performance than competing baselines, highlighting the inherent spatio-temporal capability of pretrained diffusion models for objecet tracking via segmentation. While textual inversion increases runtime, the corresponding accuracy improvement demonstrates an acceptable trade-off in practice.

\subsection{Comparison Across Diffusion Model Variants}
\label{appendix:compare_sd}
\begin{table}[h]
\centering
\caption{\textbf{Comparison of SD 1.5 and 2.1 with and without SAM on DAVIS 2017.}  
This ablation highlights the effect of backbone diffusion model and shows how performance varies with and without SAM-based refinement.}
\scalebox{0.9}{
\begin{tabular}{lcccccc}
    \toprule
    \toprule
    Model & Params & SAM & $\mathcal{J}\&\mathcal{F}_\mathrm{m}$ & $\mathcal{J}_\mathrm{m}$ & $\mathcal{F}_\mathrm{m}$ \\
    \midrule
    \multirow{2}{*}{SD 1.5} & \multirow{2}{*}{860M} & \ding{55} & 68.9 & 65.6 & 72.1 \\
                            &                        & \ding{51} & 76.4 & 74.1 & 78.8 \\
    \midrule
    \multirow{2}{*}{SD 2.1} & \multirow{2}{*}{865M} & \ding{55} & 74.8 & 70.7 & 78.9 \\
                            &                        & \ding{51} & 81.3 & 78.8 & 83.7 \\
    \bottomrule
    \bottomrule
    \end{tabular}
    }
\label{tab:ablation_model}
\end{table}
Besides our primary results based on Stable Diffusion 2.1, we also evaluate our framework using Stable Diffusion 1.5~\citep{stablediffusion} as the backbone.  
This variant processes $512 \times 512$ resolution inputs and produces latent features of size $64 \times 64$.  
We extract self-attention maps from all three attention layers in the final decoder block of the U-Net.  
All other settings remain identical to those used with the 2.1 backbone.  
As shown in \cref{tab:ablation_model}, we observe that performance trends remain consistent, confirming the general applicability of our method across diffusion model versions. Notably, the performance disparity between SD~2.1 and SD~1.5 backbones is not attributed to model size, as their U-Net parameter counts are nearly identical (865M vs.\ 860M), suggesting that other factors, such as differences in training data or procedures, play a more significant role.

\subsection{Mask Refinement with PAC-CRF}
\label{appendix:crf}
\begin{table}[ht]
\centering
\caption{\textbf{Comparison of Refinement Method on DAVIS 2017.} All models are based on SD 2.1 with DDIM inversion and textual inversion applied. Except for the refinement module, all other experimental settings are kept identical.}
\scalebox{0.9}{
\begin{tabular}{lccccc}
    \toprule
    \toprule
    Model & Refiner & $\mathcal{J}\&\mathcal{F}_\mathrm{m}$ & $\mathcal{J}_\mathrm{m}$ & $\mathcal{F}_\mathrm{m}$ \\
    \midrule
    \multirow{2}{*}{\citet{uziel2023vit}}       & \ding{55} & 74.1 & - & - \\
                                 & PAC-CRF & 76.3 & \textbf{73.8} & 78.7 \\
    \midrule
    \multirow{3}{*}{\drift{}(Ours)} & \ding{55} & 74.5 & 70.3 & 78.6 \\
                                 & PAC-CRF & \textbf{76.4} & 73.0 & \textbf{79.8} \\
                                 & \textcolor{gray}{SAM} & \textcolor{gray}{81.3} & \textcolor{gray}{78.8} & \textcolor{gray}{83.7} \\
    \bottomrule
    \bottomrule
\end{tabular}
}
\label{tab:crf}
\end{table}
In addition to our primary refinement method using SAM, we also explore the use of PAC-CRF~\citep{paccrf} as a lightweight post-processing technique for enhancing the spatial quality of predicted masks.
PAC-CRF refines a segmentation mask by enforcing local smoothness and edge-aware consistency using the underlying image as guidance.
Following prior work~\citep{paccrf, uziel2023vit}, we apply PAC-CRF with a kernel size of $5 \times 5$ and 30 refinement steps to binary masks $\hat{M} \in \{0,1\}^{H \times W}$, treating them as noisy initial labels, and use the corresponding image $I \in \mathbb{R}^{H \times W \times 3}$ to guide the refinement via pairwise potentials that penalize label inconsistencies between neighboring pixels with similar appearance.
While not as powerful as prompt-based refinement with SAM, PAC-CRF can moderately improve mask alignment near object boundaries with low computational overhead.
Quantitative results comparing refinement strategies are presented in
\cref{tab:crf}.

\section{Discussions}
\subsection{On the Role of SAM Refinement}

Zero-shot approaches inherently struggle to produce fine-grained boundaries because they never receive pixel-level supervision during training. 
Temporal propagation provides a strong semantic prior for maintaining object consistency across frames, but this prior is naturally limited to coarse localization rather than boundary-level accuracy. 
The use of SAM~\citep{sam} is therefore an optional post-hoc refinement step that adds high-resolution detail to an already localized region.
Importantly, SAM itself does not provide temporal correspondence.
Its gains appear large because once \drift{} provides a semantically correct region-of-interest, boundary sharpening can improve the $\mathcal{J}\&\mathcal{F}_\mathrm{m}$ metrics. 
However, SAM cannot correct inaccurate localizations—it only refines the given region.
Zero-shot methods can benefit from SAM, but the improvements depend on how temporally consistent the propagated regions are. 
Because \drift{} produces more temporally consistent and semantically precise regions than prior feature-similarity-based propagation methods, SAM is able to refine our results much more effectively. Consequently, \drift{} achieves the strongest performance among zero-shot approaches when paired with SAM—not because it depends on SAM, but because it provides the highest-quality propagated regions for SAM to refine.

\subsection{Limitations}
\label{appendix:limitations}
A limitation of our approach is the computational overhead of textual inversion.
This step, though required only once per object on the first frame, involves optimizing prompt embeddings and can be costly when applied to datasets with many videos or object instances.
All experiments were conducted on NVIDIA H100 94GB GPUs with the diffusion model kept frozen.
While textual inversion entails additional computation, memory usage remains modest throughout both the inversion and inference stages.
Reducing the overhead of textual inversion—through faster optimization, caching strategies, or amortized prompt learning—remains an important direction for future work.


\end{document}